\documentclass[preprint,12pt,3p]{elsarticle}

\usepackage[utf8]{inputenc}
\usepackage[T1]{fontenc}
\usepackage{lmodern}

\usepackage{amssymb,amsmath,amsfonts}
\usepackage{graphicx}
\usepackage{xcolor}
\usepackage{supertabular,enumitem}
\usepackage{multirow}
\usepackage{algorithm}
\usepackage{algorithmic}
\usepackage{listings}
\usepackage{float}
\usepackage{caption}
\usepackage{subcaption}
\usepackage{placeins}
\usepackage{booktabs}
\usepackage{url}
\usepackage{xspace}

\usepackage{siunitx}
\DeclareSIUnit{\month}{month}

\usepackage{hyperref}
\hypersetup{
    pdfauthor={Emna Boudabbous, Mohamed Karaa, Lokman Sboui, Julio Montecinos, Omar Alam},
    pdftitle={Scalable Transit Delay Prediction at City Scale},
    pdfkeywords={GTFS-Realtime, deep learning, H3 indexing, transit prediction},
    colorlinks=true,
    linkcolor=blue,
    citecolor=blue,
    urlcolor=blue
}

\providecommand{\ARXIV}{}

\begin{document}

\makeatletter
\def\ps@pprintTitle{%
 \let\@oddhead\@empty
 \let\@evenhead\@empty
 \let\@oddfoot\@empty
 \let\@evenfoot\@empty
}
\makeatother

\begin{frontmatter}

\title{Scalable Transit Delay Prediction at City Scale: A Systematic Approach with Multi-Resolution Feature Engineering and Deep Learning\tnoteref{arxiv}}

\tnotetext[arxiv]{This manuscript is a preprint of an earlier version. A revised version is currently under review.}

\author[ets]{Emna Boudabbous}  
\ead{emna.boudabbous.2@ens.etsmtl.ca}

\author[ets]{Mohamed Karaa}
\ead{mohamed.karaa.1@ens.etsmtl.ca}

\author[ets]{Lokman Sboui}
\ead{lokman.sboui@etsmtl.ca}

\author[ets]{Julio Montecinos\corref{cor1}}
\ead{julio.montecinos@etsmtl.ca}

\author[trent]{Omar Alam}
\ead{omaralam@trentu.ca}

\cortext[cor1]{Corresponding author}

\affiliation[ets]{organization={Systems Engineering Department, École de technologie supérieure (ÉTS)},
            addressline={1100 Notre-Dame Street West},
            city={Montréal},
            postcode={H3C 1K3},
            state={Québec},
            country={Canada}}

\affiliation[trent]{organization={Department of Computing and Information Systems, Trent University},
            addressline={1600 West Bank Drive},
            city={Peterborough},
            postcode={K9L 0G2},
            state={Ontario},
            country={Canada}}

\begin{abstract}
Urban bus transit agencies need reliable, network-wide delay predictions to provide accurate arrival information to passengers and support real-time operational control. Accurate predictions help passengers plan their trips, reduce waiting time, and allow operations staff to adjust headways, dispatch extra vehicles, and manage disruptions. Although real-time feeds such as GTFS-Realtime (GTFS-RT) are now widely available, most existing delay prediction systems handle only a few routes, depend on hand-crafted features, and offer little guidance on how to design a scalable, reusable architecture. 

We present a city-scale prediction pipeline that combines multi-resolution feature engineering, dimensionality reduction, and deep learning. The framework generates \num{1683} spatiotemporal features by exploring 23 aggregation combinations over H3 cells, routes, segments, and temporal patterns, and compresses them into 83 components using Adaptive PCA while preserving 95\% of the variance. To avoid the ``giant cluster'' problem that occurs when dense urban areas fall into a single H3 region, we introduce a hybrid H3+topology clustering method that yields 12 balanced route clusters (coefficient of variation 0.608) and enables efficient distributed training. 

We compare five model architectures on six months of bus operations from the Société de transport de Montréal (STM) network in Montréal. A global LSTM with cluster-aware features achieves the best trade-off between accuracy and efficiency, outperforming transformer models by 18--52\% while using $275\times$ fewer parameters. We also report multi-level evaluation at the elementary segment, segment, and trip level with walk-forward validation and latency analysis, showing that the proposed pipeline is suitable for real-time, city-scale deployment and can be reused for other networks with limited adaptation.
\end{abstract}

\ifdefined\ARXIV
\else
\begin{highlights}
    \item Developed a scalable, systematic framework for multi-resolution feature engineering, leveraging Uber's H3 spatial indexing system to generate spatial-temporal features across diverse aggregation levels.
    \item Introduced a hybrid clustering method combining H3 indexing and topological structure to mitigate the issue of disproportionately large clusters, a challenge we refer to as the ``giant cluster'' problem, which arises in dense urban mobility data.
    \item Demonstrated that a global LSTM architecture surpasses transformer-based models in trip-level delay prediction, achieving higher accuracy with significantly fewer parameters.
    \item Built a complete end-to-end prediction pipeline incorporating hierarchical feature aggregation, walk-forward validation, and thorough computational efficiency analysis.
    \item Validated the system's robustness and generalizability through extensive evaluations across model types, spatial resolutions, and challenging conditions such as extreme weather and special events.
\end{highlights}
\fi


\end{frontmatter}



\section{Introduction}
\label{sec:introduction}

Urban public transportation systems are a cornerstone of sustainable mobility in metropolitan areas, carrying millions of passengers daily and helping reduce congestion, pollution, and energy consumption~\cite{elassy2024intelligent}. For bus networks in particular, reliability is often degraded by traffic congestion, incidents, adverse weather, and operational constraints. Accurate real-time delay prediction has therefore become a key function of modern Intelligent Transportation Systems (ITS), as it directly affects service quality, passenger satisfaction, and operational efficiency~\cite{10908455}. Reliable predictions allow passengers to plan their trips better and reduce waiting time. They help agencies adjust operations, modify routes and frequencies, and improve service in ways that support public transport adoption~\cite{10908455}. As urban mobility demand grows, scalable, reliable delay-prediction systems are increasingly necessary.

Despite advances in artificial intelligence (AI) and the widespread availability of real-time data through standards such as General Transit Feed Specification Realtime (GTFS-RT), several fundamental challenges still hinder the deployment of reliable, large-scale delay prediction systems~\cite{singh2022review}. 

First, feature engineering remains largely ad hoc. Many existing studies rely on manually designed feature sets driven by researcher intuition~\cite{manual1,manual2}. Such approaches are difficult to reproduce, offer limited coverage of multi-resolution spatiotemporal patterns, and generalize poorly across different transit networks. The lack of a systematic, reproducible framework for feature generation is a central methodological gap.

Second, scalability and computational efficiency pose significant challenges for network-wide transit delay prediction. While many studies report strong performance on individual routes~\cite{9875065,petersen2019multi_output}, few address the complexities of modelling an entire metropolitan transit system comprising hundreds of routes, thousands of stops, and millions of observations~\cite{arrivalNet,9714843}. Monolithic models that treat the network as a single, undifferentiated unit are both computationally intensive and ill-suited to capturing spatial heterogeneity, particularly the operational differences between dense urban cores and lower-density suburban zones. Spatial clustering has emerged as a practical strategy to address this issue by dividing the network into subregions with similar operational profiles, thereby improving model efficiency and maintaining local delay patterns~\cite{elsa2018evaluation}. However, conventional clustering methods often rely on regular spatial grids that ignore the network topology, leading to severe data imbalances. In particular, high-density urban areas are frequently collapsed into disproportionately large clusters that dominate the dataset. We refer to this as the ``giant cluster phenomenon'': a spatial-aggregation artifact in which dense regions are represented by oversized spatial units, resulting in spatial homogenization, obscured intra-urban variability, degraded model performance, and inefficient parallel training.

Third, much of the literature focuses on experimental prototypes with limited attention to deployment-oriented evaluation. Essential aspects such as inference latency, resource usage, systematic architecture comparison, and computational efficiency are rarely reported~\cite{10908455}. This makes it difficult for practitioners to select architectures that are feasible for real-time, network-wide deployment.

Recent technological advances open new opportunities to address these gaps. Hierarchical spatial indexing systems, particularly Uber's hexagonal hierarchical spatial indexing (H3)~\cite{uber2018h3}, provide a flexible framework for multi-resolution spatial representation and allow transit patterns to be aggregated consistently at multiple scales (e.g., neighbourhood, corridor, and segment). Yet their use for transit delay prediction, especially in combination with rigorous multi-resolution feature engineering and topology-aware spatial clustering, remains relatively unexplored.

This paper presents an end-to-end pipeline for city-scale delay prediction, developed to meet the operational requirements of bus transit and to support the creation of a reusable architecture that can be readily adapted to other urban transit systems.

Our work pursues three interconnected research objectives: (1) to develop a reproducible framework for exhaustive multi-resolution spatiotemporal feature generation, (2) to design an efficient spatial clustering strategy for data organization and feature engineering that resolves the ``giant cluster problem'' while preserving network topology, and (3) to perform a rigorous comparative evaluation of multiple deep learning architectures, including computational analysis, on a complete metropolitan bus network.

The main contributions of this work are fourfold:
\begin{itemize}
    \item \textbf{Systematic multi-resolution feature engineering framework.} We introduce a reproducible framework for spatiotemporal feature generation that systematically explores aggregation combinations across H3 spatial resolutions (9 and 10), route identifiers, segments, and temporal dimensions (hour of day, time periods, and per-hour boolean intervals). The framework produces \num{1683} features capturing both local segment-level patterns and neighbourhood-level trends. We evaluate 20 dimensionality reduction methods and show that Adaptive PCA reduces this set to 83 components while retaining 95\% of the variance, making global model training computationally feasible.

    \item \textbf{Hybrid H3+topology spatial clustering for data organization.} We propose a route clustering methodology based on weighted Jaccard similarity that combines spatial coverage (H3 hexagons) with topological structure (shared segments) to organize network data and generate cluster-aware features. Using agglomerative clustering with Ward linkage, we obtain an effective configuration of 12 balanced clusters at H3 resolution 7 (hexagons of roughly $5~\text{km}^2$), after systematically evaluating more than 80 configurations. The resulting clusters (coefficient of variation $\mathrm{CV}=0.608$) support efficient feature engineering while preserving spatial and topological coherence for global model training.

    \item \textbf{Comparative evaluation of deep learning architectures.} We perform an extensive comparison of five architectures for elementary-segment bus delay prediction: LSTM (Long Short-Term Memory), XGBoost (gradient boosting trees), xLSTM (Extended LSTM), PatchTST (patch-based transformer), and Autoformer (autocorrelation transformer). All models share the same preprocessing and feature pipelines and are trained globally on the complete bus network. For each architecture, we report predictive accuracy, training time, inference latency, and memory usage. Results show that LSTM with compressed features achieves the best overall performance (elementary-level $R^2 = 0.6621$), outperforming transformer-based models by 18--52\% while maintaining favourable computational and latency profiles.

    \item \textbf{End-to-end pipeline for bus delay prediction with multi-level validation.} We implement a complete pipeline for bus delay prediction from data ingestion (GTFS static data, GTFS-RT feeds, and weather data) to distributed feature engineering, dimensionality reduction, and global model training with multi-level prediction aggregation (elementary segment $\rightarrow$ segment $\rightarrow$ trip). The system uses walk-forward temporal validation and reports RMSE, MAE, and $R^2$ at all three levels, together with runtime and resource measurements. Multi-level aggregation exhibits error cancellation: trip-level RMSE (1.85~min for the LSTM model) is substantially lower than elementary-level errors, supporting the suitability of the hierarchical framework for operational deployment.
\end{itemize}

The remainder of this paper is organized as follows. Section~\ref{sec:related_work} reviews related work in transit delay prediction, feature engineering, and distributed computing for spatiotemporal data. Section~\ref{sec:methodology} presents our methods, including the multi-resolution feature engineering framework, hybrid spatial clustering strategy, and model architectures. Section~\ref{sec:implementation_results} details the technical implementation, infrastructure choices, hyperparameter configurations, and experimental results. Section~\ref{sec:discussion} discusses implications and future research directions.

\textbf{Motivation and Problem Setting}
\label{sec:motivation}

This paper presents an end-to-end pipeline for city-scale delay prediction, developed to address the operational requirements of bus transit systems and to support the creation of a reusable architecture that can be readily adapted to other urban networks. The pipeline is demonstrated using data from the Société de transport de Montréal (STM). Control center operators must identify delayed vehicles, understand where delays accumulate, and decide when to intervene, for example, by holding vehicles, short-turning trips, or adding extra buses. At the same time, passengers rely on predicted arrival times exposed through various information channels. When these predictions are unreliable or unstable, both operators and passengers quickly lose trust in the information system.

Our goal is to provide delay predictions that are reliable at the scale of the entire STM bus network and that can be used at several spatial and temporal granularities. To achieve this, the underlying data and models are organized around a small set of basic units:

\begin{itemize}
    \item A \textbf{trip} is a single scheduled bus journey, as defined by its GTFS \texttt{trip\_id}, from its origin terminal to its destination terminal.
    \item A \textbf{segment} is a directed pair of consecutive stops $(\text{stop}_i, \text{stop}_{i+1})$ on a route. Segment travel time is computed from GPS observations using geofences around each stop~\cite{boudabbous2024analyzing}.
    \item Each segment is subdivided into fixed-length \textbf{elementary segments}. Predictions are expressed at this elementary level as \textbf{pace}, i.e., travel time per meter (seconds/meter), which normalizes for segment length and yields a more homogeneous prediction target across the network~\cite{boudabbous2024analyzing}.
\end{itemize}

These definitions ensure that delay predictions are comparable across routes of different lengths and geometries, and they provide a clear link between the model outputs and quantities of direct operational interest, such as segment and trip-level delays~\cite{boudabbous2024analyzing}.

A commonly identified requirement in the context of large urban transit systems is the need for a \emph{reusable} architecture that can generalize across networks. Building a bespoke prediction system for a single city or corridor is feasible, but maintaining it is difficult as routes change, new data sources are added, or evaluation requirements evolve. It is also hard for other agencies to reuse such a system if its components are tightly coupled to local assumptions. To address this, our design separates city-specific configuration (GTFS feed, H3 resolution, clustering parameters) from generic components (feature generation, dimensionality reduction, global modelling, and inference). The same pipeline can therefore be adapted to other bus networks by changing a limited set of configuration files and retraining the models, while preserving the overall structure.

In summary, the motivation for this work is twofold: (i) to support real operational decision making through delay predictions defined consistently at multiple spatial and temporal scales, and (ii) to do so with a scalable, reusable architecture that can be transferred across bus networks with limited engineering effort.

\section{Related Work}
\label{sec:related_work}
Public transit delay prediction has been extensively studied and remains a relevant problem, given the continuous evolution of transportation systems and their increasing complexity. In this section, we summarize pertinent prior work on delay prediction models, feature engineering methods, and challenges related to scalability and computational costs.

\subsection{Transit Delay Prediction}
\label{subsec:prediction_methods}

Early approaches to transit delay prediction relied on simple statistical assumptions, focusing primarily on historical averages and scheduled timetables. With the increasing availability of real-time sensor data and advancements in machine learning, models have gradually evolved toward data-driven and context-aware frameworks~\cite{boudabbous2024analyzing}. 

Statistical approaches include time series analysis using ARIMA~\cite{suwardo2010arima}, Kalman Filters~\cite{8691701}, mixture models~\cite{chen2025understanding}, Bayesian networks~\cite{9564537}, and Markov models~\cite{sun2025realtimebustraveltime}. Although these methods are easier to implement and interpret, and require less computational resources and data, they often struggle to capture the complex underlying patterns of traffic dynamics and to adapt to irregularities such as service disruptions and sudden events.

The proliferation of real-time transit data collected by automatic vehicle location (AVL) systems has encouraged the development of advanced deep learning models, enabling more accurate delay predictions. The models trained with GPS historical data include linear regression~\cite{6338767}, support vector machines~\cite{markovic2015analyzing}, and gradient boosting algorithms~\cite{10683692,10920146}. However, these methods require meticulous feature engineering and selection for better results.

More recently, deep learning models have become the norm for processing time-series and network data to capture complex, latent spatio-temporal patterns in road networks. Recurrent neural networks (RNNs) have become the go-to model architecture for delay prediction because they can handle complex sequential data. In~\cite{alam2021predicting}, Alam et al. built a hybrid RNN-LSTM model to predict bus arrival irregularities based on historical GPS positions and weather data for two Toronto bus service routes. Liu et al.~\cite{8954709} employed an LSTM model to predict both short and long-distance arrival times of buses to the station. Another work used BiLSTMs to estimate the bus departure time for each route, then combined them with a DNN to calculate travel time~\cite{9913942}. These works have leveraged the temporal aspect of public transit dynamics to predict delay and arrival times.

In other studies, the authors have also used spatial patterns in traffic networks via Graph Neural Networks (GNNs) to improve the modelling of the delay prediction problem. In~\cite{10.1145/3703412.3703417}, the authors combined an LSTM with a GNN that captures the static characteristics of the public transit network. This hybrid approach improved the prediction accuracy compared to using temporal and spatial features separately. Sharma et al. proposed a Temporal Graph Convolutional Network that predicts delays across a segmented road network and then aggregates them to estimate the bus arrival time at a given station~\cite{11004294}. Another study proposed a Graph Attention Network that learns dynamic correlations between bus routes and then combines them with spatio-temporal features via an attention mechanism. This significantly improves the bus delay estimation, especially for bus routes with fewer representations in the road network~\cite{10.1145/3583780.3614730}.

\subsection{Feature Engineering for Transit Delay Prediction}
\label{subsec:feature_engineering}

Feature engineering remains the cornerstone of effective predictive modelling for transit systems. While deep learning-based methods often avoid manual feature design through end-to-end learning, empirical evidence consistently shows that well-engineered features substantially improve performance, even for neural architectures.

Conventional feature engineering techniques include several categories, such as temporal, spatial, and operational (contextual) features.
Temporal features such as hours of the day, day of the week, seasonality, and holiday indicators form the foundation of most prediction models~\cite{9913942,10026635,9714843}. Effective temporal encoding requires domain knowledge to identify relevant periodicity and interaction terms. For example, rush-hour behaviour differs fundamentally between weekdays and weekends, necessitating interaction features. On the other hand, spatial information is represented as raw latitude/longitude coordinates~\cite{10.1145/3703412.3703417,10026635} or higher-level features such as regions, neighbourhoods, and street types~\cite{10.1145/3583780.3614730}. However, these representations suffer from critical limitations: coordinates lack semantic structure exploitable by tree-based models, while administrative zones exhibit arbitrary boundaries and high spatial heterogeneity. 

Operational features provide more context on bus speed, passenger load, and traffic, thereby enriching the input data~\cite{8954709,10.1145/3583780.3614730,9913942}. However, this data is often unavailable across different transit agencies. Other works have integrated weather data as input features to improve the prediction, as weather conditions affect road conditions and mobility choices~\cite{11162495}. In~\cite{alam2021predicting}, the authors improved bus arrival time prediction by 48\% by including weather data to train their LSTM model.

More recent studies have used spatial clustering to generate more granular and robust features for multiple traffic analysis and forecasting tasks~\cite{shaji2022joint,8569648}. More particularly, Uber's hexagonal spatial indexing has revolutionized the spatial partitioning techniques. H3 partitions the Earth's surface into nested hexagonal grids across 16 resolution levels (0-15) with decreasing diameter size, enabling seamless multi-resolution aggregation~\cite{uber2018h3}. Hexagons offer superior geometric properties compared to squares: uniform distance to six neighbours, minimal area distortion, and exact 1:7 parent-child subdivision ratios. At resolutions 9 (174m), 10 (66m), and 11 (25m), H3 cells naturally align with transit operational scales, namely neighbourhoods, blocks, and street segments, respectively. Multiple works have used H3 indexing for traffic analysis and travel time estimation~\cite{11004294,10.1145/3486640.3491392, 10756642}. In these works, H3-based clustering helped identify distinctive features of public transportation, such as service quality, density, and frequency.

\subsubsection{Critical Gap: Ad-Hoc Feature Selection}

A fundamental limitation across the existing literature is the \textbf{ad hoc, non-systematic nature of feature engineering}. Studies typically report 50-200 manually selected features with limited documentation of their selection criteria, making results difficult to compare and build upon~\cite{petersen2019multi_output}. This approach is inherently vulnerable to researcher bias and risks incomplete coverage of critical spatiotemporal interactions. Ultimately, these shortcomings undermine both the \textbf{reproducibility} of the findings and the \textbf{generalization} of the features to other transit networks, creating a significant barrier to progress in the field.

Despite H3's demonstrated effectiveness in mobility analytics, its systematic exploitation for \textbf{multi-resolution feature generation} in transit delay prediction remains largely unexplored. Most studies employ spatial features at a single resolution or rely on ad hoc manual selection, failing to capture the hierarchical nature of delay propagation, in which local segment-level disruptions interact with broader neighbourhood-level congestion patterns. To our knowledge, no prior work systematically explores multi-resolution feature generation across multiple H3 resolutions combined with comprehensive temporal encoding through automated aggregation combinations.

\subsection{Scalability Challenges and Distributed Computing}
\label{subsec:scalability}

Operational transit delay prediction for a metropolitan bus network requires processing millions of observations per day across hundreds of routes and thousands of stops. This volume typically exceeds the capacity of single-machine workflows, motivating the use of distributed computing frameworks.

Apache Spark is widely used for large-scale transit data processing because it offers in-memory computation, high-level DataFrame APIs, and distributed machine learning through MLlib~\cite{zaharia2018spark_guide}. For example, Lv et al.\ processed 50 million daily taxi GPS trajectories in Beijing with second-level latency using Spark clusters. Other frameworks, such as Dask and Ray, target similar workloads with different design trade-offs, emphasizing tight Python integration or low-latency task scheduling. In this work, we adopt Spark for feature engineering and model training, as it integrates with STM’s data infrastructure and can handle the scale of our datasets.

A major scalability challenge for network-wide modelling is spatial heterogeneity: delay patterns differ substantially between dense downtown corridors, suburban residential areas, and industrial zones. Treating the entire network as a single global context leads to (i) computational intractability on multi-million-row datasets with thousands of features, (ii) limited ability to learn location-specific patterns because local signals are drowned in global noise, and (iii) prohibitive training times. Spatial partitioning addresses these issues by decomposing the network into geographic clusters and training specialized models in parallel. However, naive partitioning using regular grids or coarse H3 cells can lead to severe data imbalance. Dense urban cores form a single, large cluster that contains most observations, while peripheral regions form many small clusters with sparse data. We refer to this as the \emph{``giant cluster problem''}. Our empirical analysis in Section~\ref{subsec:h3_topology_clustering} shows that naive H3 partitioning at resolution~8 yields a coefficient of variation (CV) above 2.0 and imbalance ratios above 40$\times$, whereas our hybrid H3+topology approach achieves CV~=~0.608 and an imbalance ratio of 1.90$\times$.

Topology-aware clustering strategies for transit prediction remain limited. Existing spatial clustering methods typically rely only on geographic proximity, ignoring transit-specific structure such as shared route segments and operational patterns. As a result, geographically close but operationally distinct segments may be grouped, while segments that share the same routes may be split across clusters. To our knowledge, no prior work combines H3-based spatial partitioning with an explicit topological similarity measure to improve cluster balance and operational coherence for distributed training jointly. This gap motivates the hybrid H3+topology clustering method introduced in Section~\ref{subsec:h3_topology_clustering}.

Finally, production systems must balance prediction accuracy, inference latency, and resource consumption. For real-time passenger information, latency budgets are often on the order of tens of milliseconds, which can make highly complex models impractical even if they are more accurate. Techniques such as model distillation, feature pruning, quantization, and cascaded architectures have been proposed to manage these trade-offs. Petersen et al.\ explicitly document this tension: an LSTM model attains lower RMSE than linear regression but at the cost of an order-of-magnitude higher latency; in strict real-time settings, the simpler model may therefore be preferred despite a 26\% accuracy loss~\cite{petersen2019multi_output}. In our evaluation, consequently, we report not only accuracy metrics (RMSE, MAE, $R^2$) but also training time and inference latency to assess whether candidate architectures are feasible for deployment.

\subsection{Validation Methodologies and Production Deployment}
\label{subsec:validation}

\subsubsection{Evaluation Protocols}

Most transit delay prediction studies employ some form of cross-validation, but temporal dependencies require specialized protocols. Walk-forward (rolling-origin) validation has been recognized as more appropriate for time series data because it respects causal ordering and avoids leaking future information into the training set~\cite{vlahogianni2014short_term_review}. Nevertheless, many works still rely on random train--test splits, which can overestimate performance and complicate comparisons across studies. Reported metrics also vary considerably: RMSE, MAE, and MAPE are most common, with occasional use of $R^2$ and classification-style metrics (e.g., on-time vs.\ late). This heterogeneity across protocols and metrics limits the ability to compare models fairly and assess whether a given approach is suitable for deployment.

\subsubsection{Production Deployment and Reusable Architectures}

Beyond offline evaluation, a smaller body of work considers deployment aspects of data-driven transit prediction systems. Several papers describe prototype real-time arrival prediction services integrated with passenger information systems or control centers, often using microservice-based backends or streaming frameworks for ingesting AVL and GTFS-RT data (e.g.,~\cite{petersen2019multi_output,arrivalNet}). These systems typically emphasize latency and robustness requirements and sometimes discuss monitoring and retraining strategies. More general ITS frameworks also propose modular architectures for processing streams of transportation data and serving predictions to multiple applications~\cite{zaharia2018spark_guide,vlahogianni2014short_term_review}.

However, most of these deployments are tailored to a single city, operator, or corridor and do not aim to provide a reusable, network-agnostic architecture. The feature engineering pipelines, data organization strategies, and model-selection procedures are often tightly coupled to local assumptions and rarely documented at a level that would allow other agencies to adopt them directly. In particular, we find little work that (i) treats feature generation, spatial clustering, dimensionality reduction, and model training as components of a configurable architecture, and (ii) evaluates this architecture explicitly in terms of both predictive performance and computational characteristics (training time, inference latency, and resource usage).

This gap motivates the system-level focus of our work. Rather than proposing yet another isolated prediction model, we aim to design and evaluate an end-to-end, reusable architecture for large-scale bus delay prediction that can be adapted to different networks with limited configuration changes.


\subsection{Summary and Research Gaps}
\label{subsec:summary_gaps}

Our review of the literature highlights four research gaps that motivate this work.

\textbf{Gap 1: Non-systematic feature engineering.}
Most existing studies rely on ad-hoc, manually designed feature sets, typically containing 50--200 features~\cite{petersen2019multi_output}. The researcher's intuition often drives feature selection and is only briefly documented, limiting reproducibility and transferability. While some recent works use H3 for spatial indexing~\cite{11004294,10756642}, the systematic use of multi-resolution spatial features remains limited. We found no study that combines multi-resolution H3 aggregation (e.g., resolutions 9 and 10) with a structured exploration of temporal encodings through automated aggregation combinations.

\textbf{Gap 2: Unresolved scalability challenges.}
Many contributions target a single route or a small subset of the network~\cite{9875065,petersen2019multi_output}. The few network-wide studies that exist either train a single monolithic global model (leading to prohibitive training times and spatial dilution of local patterns) or apply naive spatial partitioning schemes that create severe data imbalance between clusters. In particular, we did not find prior work that combines H3-based geographic partitioning with an explicit topological similarity measure to obtain balanced spatial clusters and to address the ``giant cluster problem'' observed in dense urban cores.

\textbf{Gap 3: Limited guidance on operational and reusable architectures.}
Most papers present experimental prototypes with little discussion of how the proposed methods could be embedded in an operational system. Production-related aspects such as inference latency, resource constraints, monitoring, model retraining, and integration with existing transit management tools are rarely covered. Moreover, architectures are typically described for a single city or corridor, with tightly coupled preprocessing and modelling components, which makes reuse by other agencies difficult. There is a lack of work that treats feature generation, spatial clustering, dimensionality reduction, and model training as configurable components of a reusable architecture for large-scale bus delay prediction.

\textbf{Gap 4: Insufficient validation rigour.}
Many studies employ simple temporal holdout splits without walk-forward cross-validation, limited hyperparameter optimization, and few ablation studies. As a result, it is often unclear which feature families or model components drive performance gains. Long-term evaluation, robustness under atypical conditions (e.g., disruptions or extreme weather), and detailed reporting of experimental protocols are also uncommon, making it hard to compare approaches or assess their suitability for deployment.
\section{Methodology}
\label{sec:methodology}

\begin{figure}[htbp]
\centering
\includegraphics[width=\textwidth]{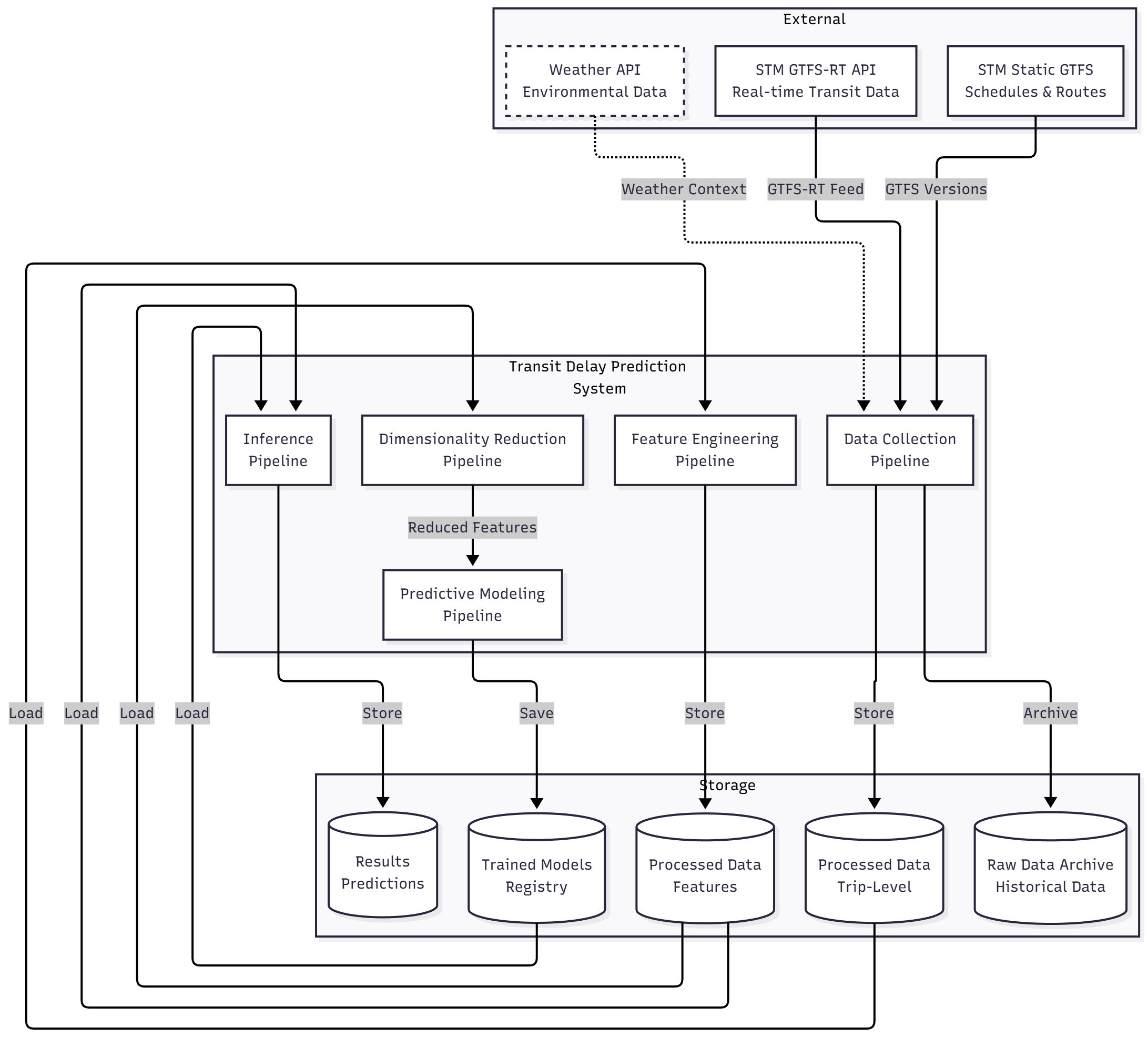}
\caption{Overview of the transit delay prediction architecture, showing the five main pipelines and their interactions with external data sources and data storage.}
\label{fig:system_context}
\end{figure}

This section describes the architecture and methods used for bus delay prediction.  
Our system is organized as five interconnected pipelines: (i) data collection,  
(ii) feature engineering, (iii) dimensionality reduction, (iv) predictive modelling,  
and (v) inference. Figure~\ref{fig:system_context} provides an overview of these  
components and their interactions with external data sources and data stores.

\subsection{Data Collection Pipeline}
\label{subsec:data_collection_pipeline}

\begin{figure}[htbp]
\centering
\includegraphics[width=0.6\textwidth]{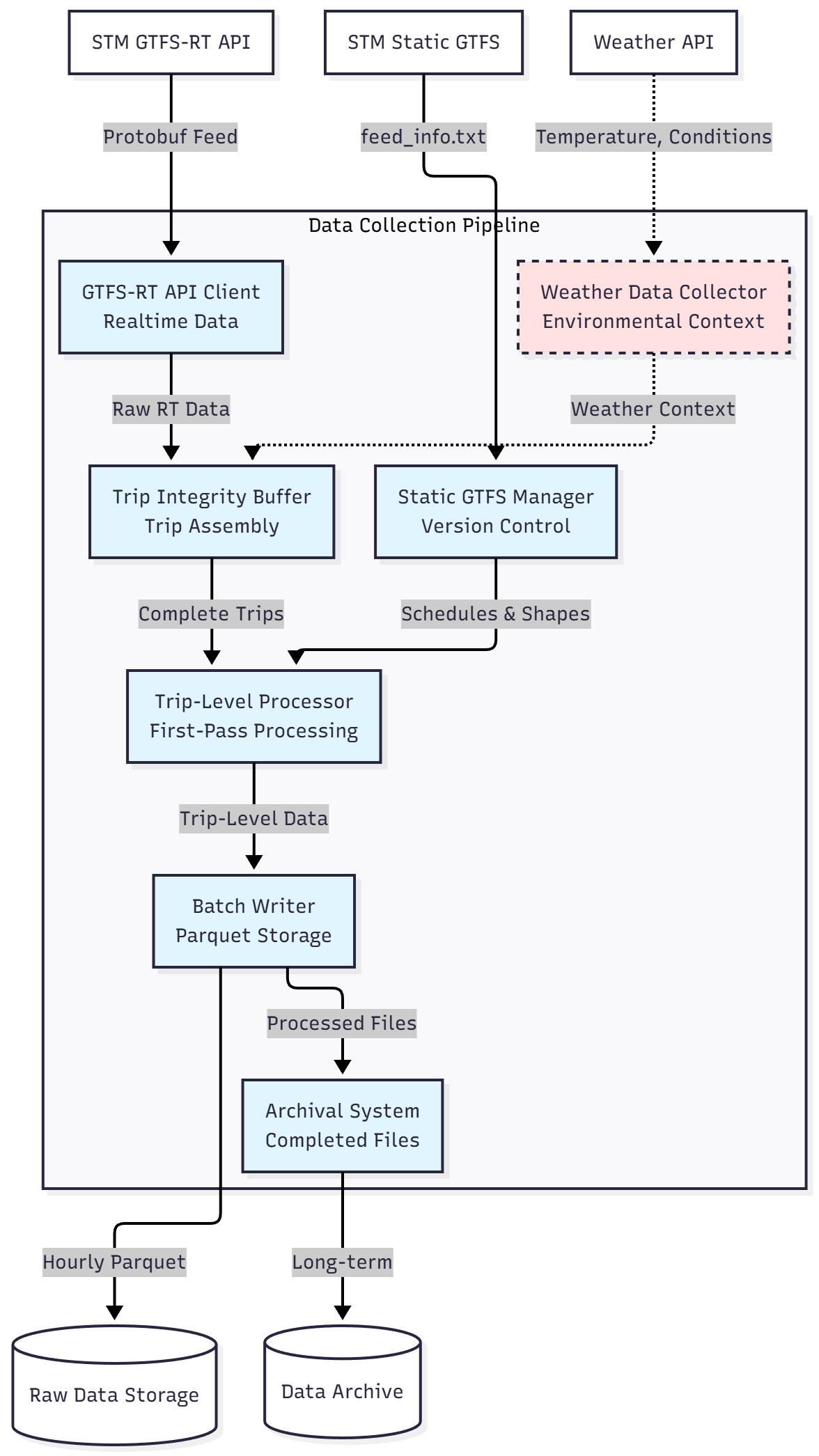}
\caption{Data collection pipeline, from raw GTFS/GTFS-RT and weather feeds to validated trip-level records.}
\label{fig:collection_pipeline}
\end{figure}

The data collection pipeline converts heterogeneous transit data sources into
trip-level records suitable for machine-learning models (Figure~\ref{fig:collection_pipeline}).  
GTFS static schedules provide the network topology (stops, routes, and stop
sequences), while GTFS-RT vehicle position updates supply real-time observations.
We align these sources in time, associate vehicle positions with scheduled trips,
and enrich each record with city-wide weather information. Schedule updates are
handled by storing successive GTFS snapshots together with their validity
periods and linking observations to the corresponding schedule version. Details
of collection frequency and aggregation procedures are provided in
Section~\ref{subsubsec:collection_impl}.

\subsubsection{Trip-level processing}
\label{subsubsec:trip_processing}

We define a \emph{trip} as a single scheduled bus journey from its origin terminal
to its destination terminal, identified by its GTFS \texttt{trip\_id}. The goal of the trip-level processing step is to transform the raw stream of GTFS-RT vehicle
positions into one record per realized trip with consistent timestamps and delay
values.

Starting from the GTFS static feed and the service calendar, we first expand all
scheduled trips for each service day. GTFS-RT vehicle positions are then matched
to these trips planned using the vehicle identifier, route, direction, and
timestamp. For each matched trip, we build an ordered sequence of observations
along the stop sequence and estimate arrival times at stops using geofences
around stop locations. Delay at a stop is computed as the difference between the
observed and scheduled arrival times.

We apply a set of rule-based quality checks to filter unreliable records. Typical
checks include discarding trips with too few observations, non-monotonic
timestamps, unrealistically high speeds between consecutive points, or locations
that deviate excessively from the scheduled route. Records that fail these
checks are excluded from subsequent stages. Weather information for the
corresponding time interval and area is then joined to each valid trip record.
Implementation details for the matching procedure, alignment windows, and
filtering thresholds are provided in
Section~\ref{subsubsec:collection_impl}.

\subsection{Feature Engineering Pipeline}
\label{subsec:feature_engineering_pipeline}

\begin{figure}[htbp]
\centering
\includegraphics[width=0.55\textwidth]{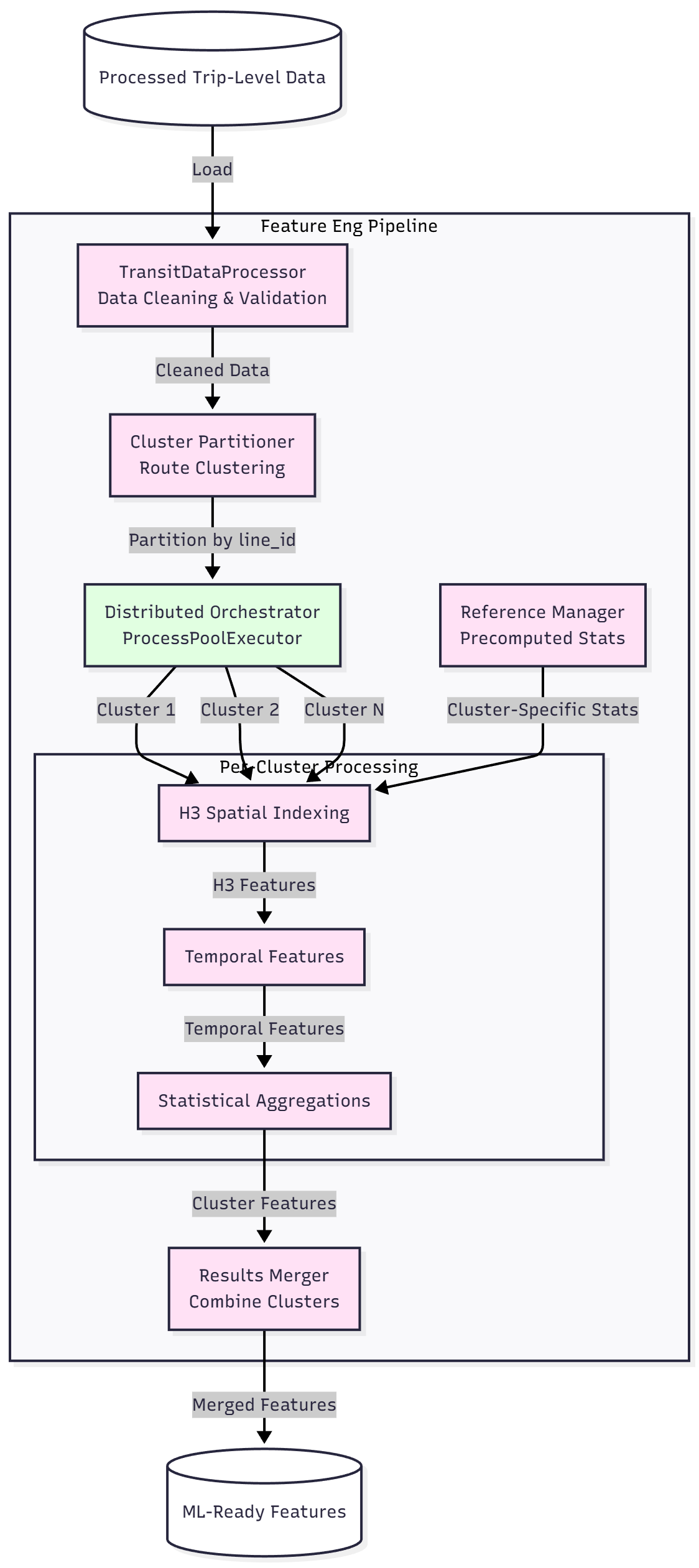}
\caption{Feature engineering pipeline, from trip-level records to a spatiotemporal feature matrix for modelling.}
\label{fig:feature_engineering_pipeline}
\end{figure}

The feature engineering pipeline transforms trip-level records into a feature matrix for
modelling (Figure~\ref{fig:feature_engineering_pipeline}). Instead of choosing features
manually, we use a fixed procedure that applies the same set of aggregation operations for
every experiment, so that the resulting feature set is well defined and easy to reproduce.

As introduced in Section~\ref{subsubsec:trip_processing}, each realized trip is represented
as an ordered sequence of segments, where a segment is the directed pair of consecutive
stops on a route. For modelling, we further subdivide each segment into fixed-length
\emph{elementary segments} and define the prediction target as elementary-segment
\emph{pace} (seconds per meter). Pace is less sensitive to segment length than raw travel
time and provides a more homogeneous target across the network.

On top of this target, we construct spatiotemporal features by aggregating historical
values over different grouping strategies. The aggregation framework combines spatial
groupings (H3 cells at resolutions 9 and 10, route and segment identifiers), temporal
groupings (hour of day, broader time-of-day periods, and per-hour indicator windows),
and route-based groupings. In total, 23 distinct combinations of spatial and temporal
groupings are considered. For each combination, we compute a small set of summary
statistics (mean, standard deviation, minimum, maximum, selected quantiles, count, and
sum) over historical delay or pace values. Together, these operations yield 1\,683
features per elementary segment, spanning both local segment-level patterns and
neighbourhood-scale trends.

The implementation of the aggregation framework, including the exact grouping definitions and a list of statistical functions, is described in
Section~\ref{subsubsec:feature_engineering_impl}. Summary statistics on feature counts
and the data volume after feature engineering are reported in
Section~\ref{subsubsec:experimental_setup_overview}.

\subsection{Dimensionality Reduction Pipeline}
\label{subsec:dimensionality_reduction_pipeline}

\begin{figure}[htbp]
\centering
\includegraphics[width=0.4\textwidth]{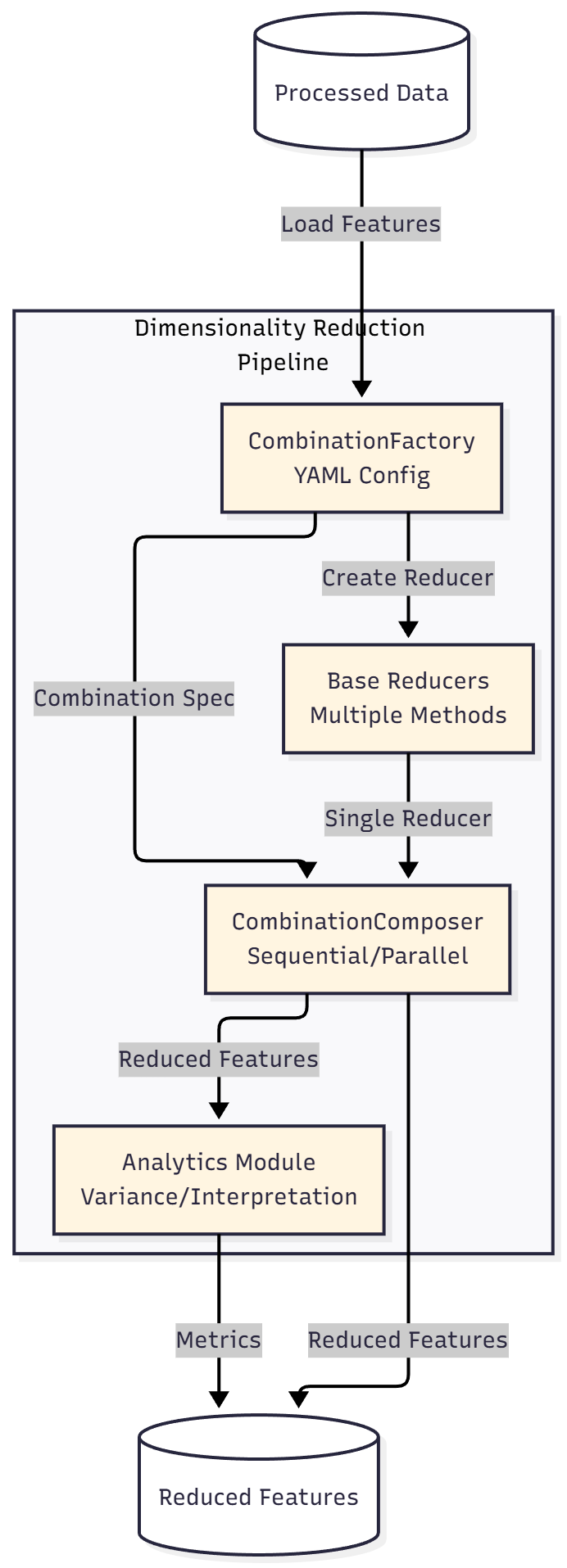}
\caption{Dimensionality reduction pipeline, from high-dimensional engineered features to a compact set of components used by all models.}
\label{fig:dr_pipeline}
\end{figure}

The dimensionality reduction pipeline maps the \num{1683} engineered features
(Section~\ref{subsec:feature_engineering_pipeline}) to a lower-dimensional
representation shared by all models (Figure~\ref{fig:dr_pipeline}). Reducing the feature space is necessary to keep training and inference times manageable while
preserving the information needed for accurate predictions.

Instead of adopting a single method by default, we evaluated a set of 20
dimensionality reduction techniques, following the comparative protocol of
Sadegh-Zadeh et al.~\cite{sadegzadeh2024comparative}. The candidate methods
include linear techniques (PCA and its variants), supervised projections
(e.g.\ LDA), nonlinear manifold methods (e.g.\ UMAP), and two-stage
compositions. For each method we measured (i) downstream prediction performance
of a reference LSTM model, (ii) training time and memory usage, and
(iii) the number of components required to reach a given level of explained
variance. This evaluation focuses on methods that can be trained on our
full-scale feature matrix and applied efficiently at inference time.

Based on these results, we selected Adaptive PCA as the default dimensionality
reduction method for the rest of the paper. In our setting, it compresses the
1\,683 features to 83 components while retaining 95\% of the variance and
achieving the best overall trade-off between accuracy and computational cost.
All subsequent experiments use these 83 components as inputs to the predictive
models. Detailed evaluation results and ablation studies for the candidate
methods are presented in Section~\ref{subsubsec:dr_impl} and
Section~\ref{subsec:experimental_results}.

\subsection{Predictive Modeling Pipeline}
\label{subsec:predictive_modeling_pipeline}

\begin{figure}[htbp]
\centering
\includegraphics[width=0.99\textwidth]{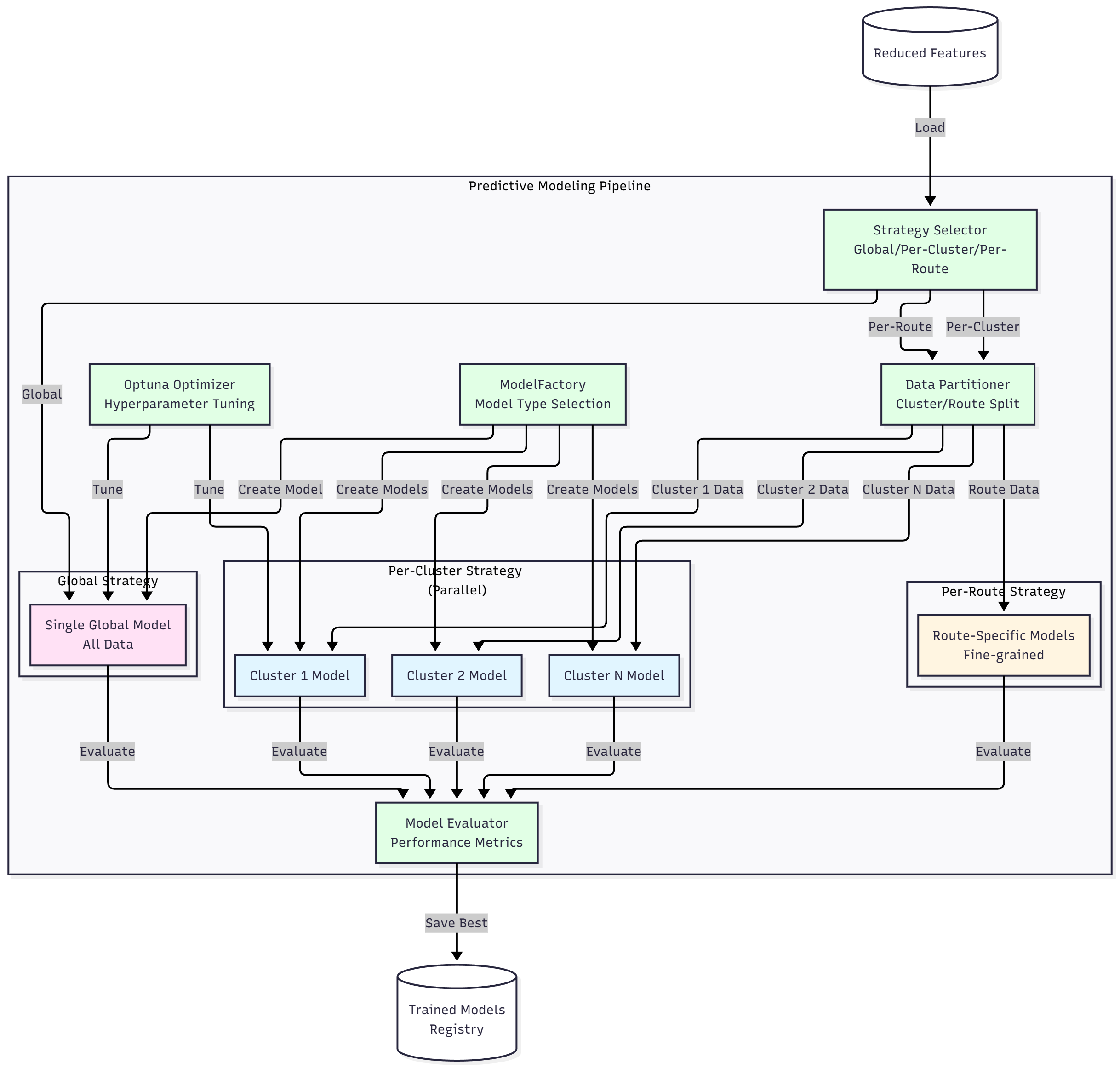}
\caption{Predictive modelling pipeline, showing global model training on reduced features with cluster-aware inputs.}
\label{fig:modeling_pipeline}
\end{figure}

The predictive modelling pipeline trains delay prediction models on the
dimensionally-reduced feature matrix produced in
Section~\ref{subsec:dimensionality_reduction_pipeline}. We adopt a global
modelling strategy in which a single network-wide model is trained on the
complete dataset, using the 83 Adaptive PCA components together with additional
categorical features such as cluster identifiers. Cluster IDs allow the model to
learn cluster-specific delay patterns while keeping a unified architecture.
This approach avoids the operational overhead of maintaining separate models
per cluster and is compatible with recent practices in large-scale traffic
forecasting~\cite{sc24supercomputing}. The overall
structure of the pipeline is illustrated in
Figure~\ref{fig:modeling_pipeline}.

We evaluate five model architectures, described in
Section~\ref{subsec:predictive_models}: LSTM, xLSTM, XGBoost, PatchTST, and
Autoformer. All models share the same input representation and prediction
target, namely, elementary-segment pace (seconds per meter), as defined in
Section~\ref{subsec:feature_engineering_pipeline}. Training uses a walk-forward
temporal cross-validation (Section~\ref{subsubsec:temporal_cv}) to respect
causality and to obtain realistic performance estimates for deployment. For each
architecture, we report RMSE, MAE, and $R^2$ at the elementary-segment, segment, and
trip level, as well as training time and inference latency. The comparative
results of this evaluation are presented in
Section~\ref{subsec:experimental_results}.

\subsection{Inference Pipeline}
\label{subsec:inference_pipeline}

\begin{figure}[htbp]
\centering
\includegraphics[width=0.5\textwidth]{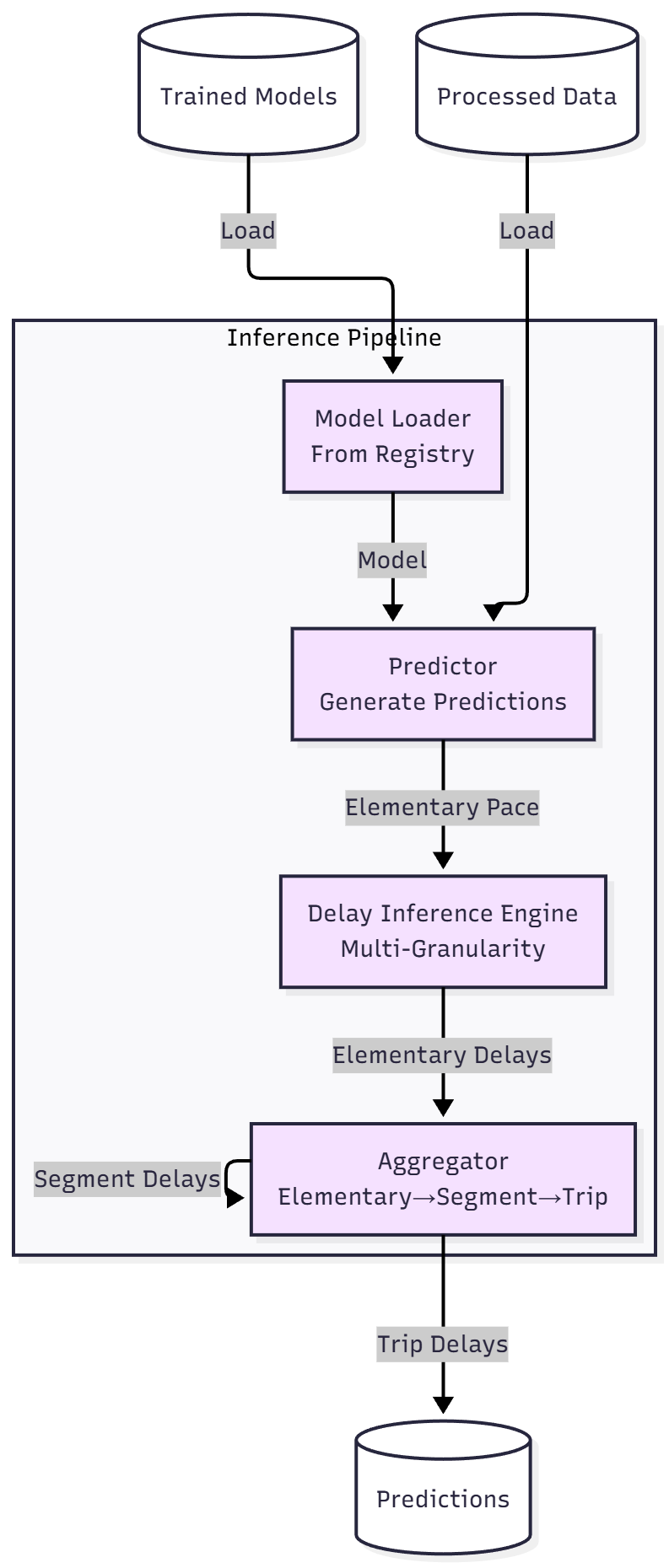}
\caption{Inference Pipeline showing delay aggregation hierarchy from elementary segments to trip-level predictions}
\label{fig:inference_pipeline}
\end{figure}

The Inference Pipeline (Figure~\ref{fig:inference_pipeline}) transforms model predictions into actionable delay estimates suitable for real-time operational use. Converting predicted elementary segment pace into delay estimates is essential because pace predictions alone are not directly actionable for transit operations—operators and passengers need delay information relative to scheduled times. Hierarchical aggregation is necessary because transit operations require predictions at multiple granularities: elementary segment predictions enable stop-level adjustments, segment predictions support route planning decisions, and trip-level predictions enable schedule adherence monitoring and passenger information systems. Multi-level output is critical because different operational use cases require predictions at various aggregation levels, as justified in the evaluation protocol (Section~\ref{subsubsec:multilevel_evaluation}). The implementation details of the inference pipeline, including pace-to-delay conversion, aggregation hierarchy, and performance optimizations, are described in Section~\ref{subsubsec:inference_impl}. The resulting inference latency and error cancellation outcomes are presented in Section~\ref{subsec:experimental_results}.

\subsection{Hybrid H3+Topology Spatial Clustering Strategy}
\label{subsec:h3_topology_clustering}

\begin{figure}[htbp]
\centering
\includegraphics[width=\textwidth]{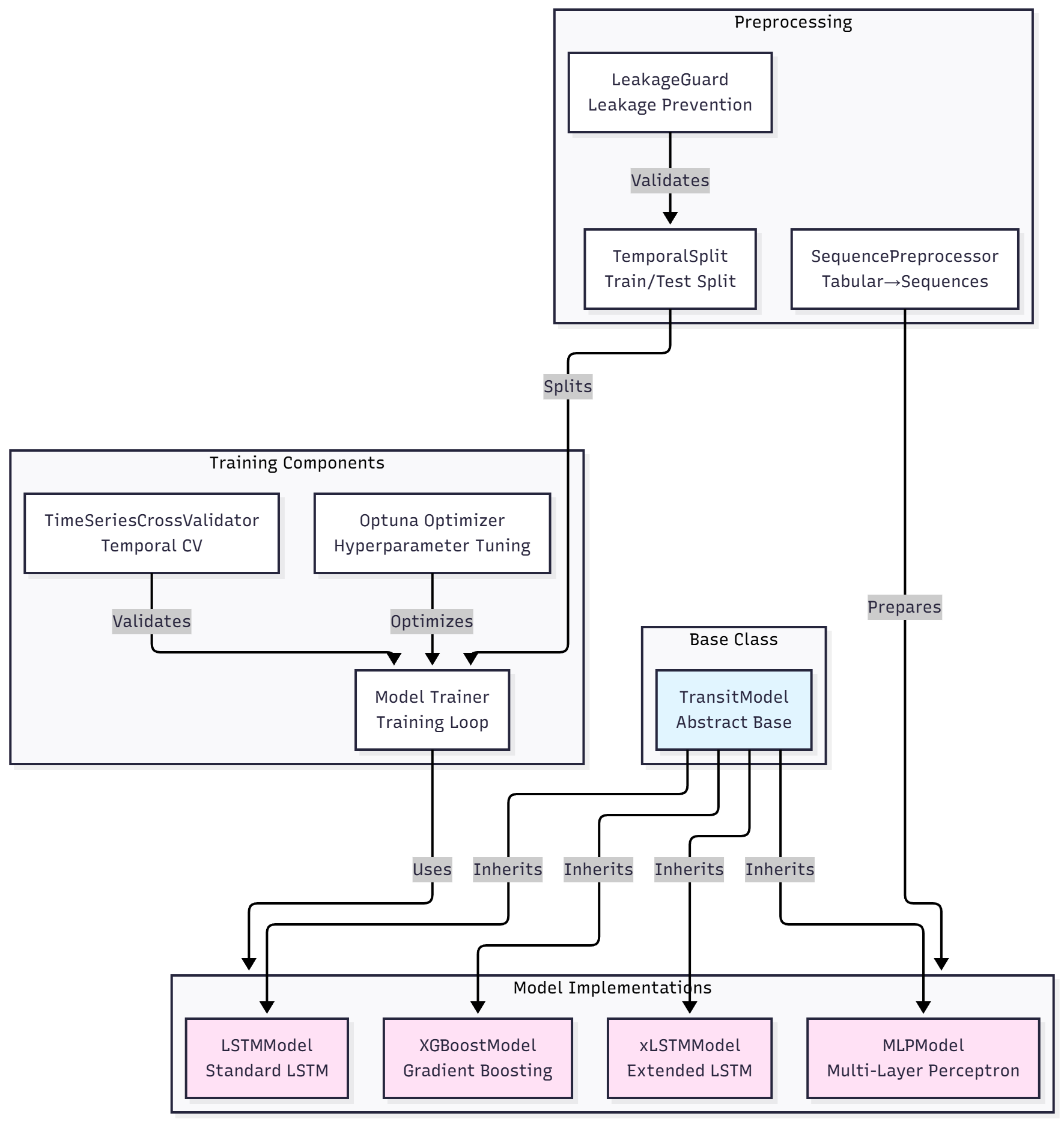}
\caption{Route clustering system components and hybrid H3+topology similarity.
Routes are grouped into clusters using a combination of spatial overlap in H3
hexagons and shared route segments.}
\label{fig:clustering_components}
\end{figure}

A critical challenge in scaling transit prediction is what we term the
\emph{``giant cluster'' problem''}: naive geographic partitioning can produce one or
two huge clusters that contain 60--80\% of the data, while the peripheral areas remain data-sparse. This imbalance undermines distributed processing and
leads to unstable models for regions with little data. To reduce this effect, we use a hybrid clustering strategy that combines H3-based spatial coverage with
network topology information, as illustrated in
Figure~\ref{fig:clustering_components}.
For routes $r_i$ and $r_j$ we define a combined similarity
\begin{equation}
\label{eq:combined_similarity}
\mathrm{similarity}_{\mathrm{combined}}(r_i, r_j)
= w_{\mathrm{spatial}} \, J_{\mathrm{H3}}(r_i, r_j)
+ (1 - w_{\mathrm{spatial}}) \, J_{\mathrm{segment}}(r_i, r_j),
\end{equation}
where $J_{\mathrm{H3}}$ quantifies spatial overlap via H3 hexagons,
$J_{\mathrm{segment}}$ measures the proportion of shared route segments, and
$w_{\mathrm{spatial}} \in [0,1]$ balances the two contributions. The intuition is that routes which operate in similar areas and share many segments should be
clustered together so that each cluster corresponds to a coherent group of
routes with similar operating conditions.

The clustering procedure follows three stages. First, we establish geographic
coherence by assigning all observations to H3 hexagons and grouping routes
according to their spatial footprints. We evaluate several candidate H3
resolutions (6, 7, and 8); resolution~7 (hexagons of roughly 5~km$^2$) offers a
good compromise between spatial detail and data volume. This initial step
reveals giant clusters where dense urban cores dominate the distribution.

Second, we refine the large clusters using topology-aware analysis. For each giant cluster, we construct a route connectivity graph in which edges reflect
$J_{\mathrm{segment}}$ similarity, and we apply community detection to split
the cluster into subgroups that preserve route relationships while improving
balance.

Third, we perform a simple rebalancing step: clusters that are too small are
merged with neighbouring clusters that have similar spatial and topological
profiles, while clusters that remain too large are recursively subdivided. The resulting configuration achieves substantially more uniform cluster sizes while
maintaining spatial and topological coherence. Algorithmic details of the three
stages are given in Section~\ref{subsubsec:clustering_algorithms}, and
quantitative results on cluster balance and prediction performance are reported
in Section~\ref{subsubsec:clustering_results}.

\subsection{Hybrid Parallelization Strategy}
\label{subsec:hybrid_parallelization}

The system combines process-based parallelization (CPU-bound tasks: model training with isolated memory) and thread-based concurrency (I/O operations: data loading, model serialization). Memory optimization via lazy loading, selective column reading, and feature streaming ensures scalability. Fault tolerance through checkpoint-based recovery and idempotent operations enables reliable distributed training. Implementation details in Section~\ref{subsec:tech_stack}.

\subsection{Predictive Models and Architectures}
\label{subsec:predictive_models}

To compare different modelling approaches within the proposed framework, we evaluate five architectures that represent three major families of time-series
models: XGBoost (gradient boosting trees), LSTM (recurrent network), xLSTM
(extended LSTM), PatchTST (patch-based transformer), and Autoformer
(autocorrelation transformer). This selection allows us to contrast tree-based,
recurrent, and transformer-based methods under a common preprocessing and
evaluation setup.

XGBoost is included as a strong tree-based baseline. It has been widely used for
structured spatiotemporal tabular data, where categorical identifiers (e.g.,
routes, time periods) interact with continuous statistics, such as delays or
pace~\cite{xue2025dmtngnn}. Tree-based models naturally handle mixed feature types
and can capture nonlinear interactions without additional feature engineering,
which is attractive for the feature-rich setting considered here.

LSTM architectures are used to capture sequential temporal dependencies in delay
patterns along a route. Recurrent networks are well-suited to modelling
stop-to-stop delay propagation, where conditions at one stop influence the next
stops in the sequence. Compared with transformer-based models, LSTMs typically
require fewer parameters for a given input size, which makes them a natural
candidate for our compressed feature space
(Section~\ref{subsubsec:experimental_setup_overview}).

xLSTM extends standard LSTM cells with additional gating and memory mechanisms
designed to represent longer-range temporal relationships. We include xLSTM to
test whether these extensions help capture delay propagation across extended
time windows, such as interactions between morning peak traffic and afternoon
operations.

PatchTST and Autoformer are transformer-based architectures that have shown
competitive performance on time-series forecasting benchmarks. PatchTST groups
input sequences into patches before applying self-attention, which reduces
computational cost while preserving local temporal structure. Autoformer
replaces standard attention with an autocorrelation mechanism coupled with
series decomposition, aiming to model seasonal and periodic patterns more
explicitly. Evaluating these architectures allows us to assess the benefits and
costs of transformer-style models for large-scale bus delay prediction.

All models are trained globally on the complete network dataset using the same
input representation (Adaptive PCA components and auxiliary features) and the
same training protocol, including walk-forward temporal validation
(Section~\ref{subsubsec:temporal_cv}). This common setup ensures a fair
comparison across architectures. Quantitative results and a discussion of
accuracy–efficiency trade-offs are presented in
Section~\ref{subsec:experimental_results}.

\subsubsection{Hyperparameter Optimization Strategy}
\label{subsubsec:hyperparameter_optimization}

Production deployment requires balancing multiple objectives beyond prediction accuracy alone. Inference latency and memory footprint constrain real-world deployment, particularly for real-time transit systems where sub-second response times are essential. We employ Bayesian optimization with a composite objective function that balances accuracy (70\%), latency (20\%), and memory footprint (10\%), reflecting practical deployment priorities where accuracy matters most but operational constraints cannot be ignored~\cite{sc24supercomputing}. This multi-objective approach ensures that models meeting accuracy targets also satisfy operational constraints essential for production deployment, following recent best practices in deep learning optimization~\cite{kartini2025dimensionality}. This optimization strategy is applied during model development to select optimal hyperparameters for each architecture, enabling fair comparison while ensuring production viability.

\subsection{Evaluation Protocol}
\label{subsec:evaluation_protocol}

The evaluation protocol is designed to provide realistic performance estimates
for deployment and to support fair comparisons between models and clustering
strategies. It combines walk-forward temporal cross-validation,
multi-level evaluation at different aggregation scales, feature-importance
analysis, and statistical testing. Implementation details are given in
Sections~\ref{subsubsec:temporal_cv_impl},
\ref{subsubsec:feature_importance_impl}, and
\ref{subsubsec:model_comparison_impl}.

\subsubsection{Walk-Forward Temporal Cross-Validation}
\label{subsubsec:temporal_cv}

Standard $k$-fold cross-validation is not appropriate for time-series data
because it can leak future information into the training set. We therefore use
walk-forward temporal cross-validation. The dataset described in
Section~\ref{subsubsec:experimental_setup_overview} is split into five
sequential folds, each corresponding to a contiguous time period. For each fold,
models are trained on past data and evaluated on a subsequent validation and
test window, with an explicit temporal gap between training and test periods to
reduce leakage. This rolling-window design respects chronological ordering and
yields performance estimates that are consistent with the intended deployment
scenario. The detailed procedure for computing time boundaries, performing data
splits, and enforcing temporal gaps is described in
Section~\ref{subsubsec:temporal_cv_impl}.

\subsubsection{Multi-Level Evaluation Strategy}
\label{subsubsec:multilevel_evaluation}

Transit operators use predictions at several levels of aggregation, from local
segments to complete trips. We therefore report metrics at three levels: elementary segment, segment, and trip (see Section~\ref{subsec:inference_pipeline} for the corresponding operational use cases). Evaluating all three provides complementary information: elementary
metrics reflect local accuracy, segment metrics capture line-level behaviour,
and trip metrics reveal how errors accumulate or cancel along a whole journey.
The latter is significant for passenger-facing applications, where trip-level delay is the primary quantity of interest.

\subsubsection{Feature Importance and Model Comparison}
\label{subsubsec:feature_ablation}

To assess whether the feature engineering framework captures meaningful
spatiotemporal patterns, we perform model-agnostic feature-importance analysis
using SHAP (SHapley Additive exPlanations). SHAP values are computed for the
best-performing models and aggregated to obtain (i) importance scores for
individual features and (ii) group-level scores for feature families such as H3
resolutions, route identifiers, and temporal encodings. We also conduct
progressive feature ablation experiments to examine how performance changes as
feature subsets are removed. Stability of importance rankings across
cross-validation folds is used as an additional robustness check. The
computational procedure is detailed in
Algorithm~\ref{alg:feature_importance_summarized} in
Section~\ref{subsubsec:feature_importance_impl}.

For comparative evaluation of models and clustering strategies, we train all
combinations under the same walk-forward protocol and apply statistical tests to
their performance metrics. After checking distributional assumptions, we use
appropriate parametric or non-parametric tests with multiple comparisons. We apply corrections to control the family-wise error rate. The whole procedure for model
training, metric aggregation, and hypothesis testing is summarised in
Section~\ref{subsubsec:model_comparison_impl}.

\section{Implementation and Results}
\label{sec:implementation_results}

\subsection{Technology Stack}
\label{subsec:tech_stack}

Python implementation with Apache Parquet storage (5:1 compression, columnar layout for \num{1600}+ features), hierarchical partitioning by \texttt{route\_id}/\texttt{date} (99\% storage overhead reduction). Hybrid parallelization: \texttt{ProcessPoolExecutor} for CPU tasks ($N_{\text{workers}} = N_{\text{cores}} - 1$), \texttt{ThreadPoolExecutor} for I/O. Memory optimization via lazy loading and Apache Arrow memory-mapping. H3 geospatial indexing (\texttt{H3-py}).

\subsection{Clustering Implementation Details}
\label{subsec:clustering_implementation}

This subsection details the specific parameter configurations, optimization process, and resulting cluster characteristics that enable efficient spatial partitioning of the transit network.

\subsubsection{Optimal Clustering Configuration}
\label{subsubsec:optimal_clustering_config}

The final clustering configuration was determined through an extensive hyperparameter search that combined a systematic grid search and Bayesian optimization. The grid search explored over 80 configurations spanning H3 resolutions (6, 7, 8), cluster counts (8, 10, 12, 15, 20), spatial weights (0.3, 0.5, 0.7), and linkage methods (Ward, Complete, Average). For each configuration, we computed cluster quality metrics including coefficient of variation (CV), imbalance ratio, silhouette score, and Davies-Bouldin index.

Following the grid search, Bayesian optimization using Optuna refined the continuous parameters (spatial weight, distance threshold) over 50 additional trials. The optimization objective balanced cluster variance (minimize CV) with spatial coherence (maximize silhouette score) and topological similarity (minimize inter-cluster route overlap). This multi-objective optimization employed a weighted scalarization approach with empirically tuned weights: 0.5 for CV, 0.3 for silhouette, and 0.2 for topological coherence. The optimization procedure evaluates each configuration by computing cluster quality metrics and selecting the configuration that optimizes the composite objective function. The resulting optimal configuration and performance metrics are presented in Section~\ref{subsubsec:clustering_results}.

\subsubsection{Three-Stage Clustering Algorithm Implementation}
\label{subsubsec:clustering_algorithms}

The three-stage clustering methodology (Section~\ref{subsec:h3_topology_clustering}) is implemented as follows. These algorithms translate the conceptual framework into executable steps that systematically resolve the ``giant cluster problem''. The H3 resolution parameter $r$ is determined through systematic grid search (Section~\ref{subsubsec:optimal_clustering_config}), with resolution 7 identified as optimal for our network. The algorithms below use this optimal resolution value.

\paragraph{Stage 1: Coarse Geographic Clustering}
The initial stage establishes geographic coherence using H3 resolution $r$ (optimal value: $r=7$, ~5 km² hexagons):
\begin{algorithm}[h]
\caption{Stage 1: Coarse Geographic Clustering}
\label{alg:stage1_geographic}
\footnotesize
\setlength{\itemsep}{-0.5ex}
\begin{algorithmic}[1]

\REQUIRE Transit observations with GPS coordinates $(\text{lat}, \text{lon})$; H3 resolution $r$ (recommended: $r = 7$)
\ENSURE Geographic cluster assignments $\text{cluster}_{\text{geo}}$; giant cluster list $\text{giant}_{\text{clusters}}$

\STATE \textbf{H3 Index Assignment}
\FOR{each observation $\text{obs}_i$ with coordinates $(\text{lat}_i, \text{lon}_i)$}
    \STATE Compute H3 index: $h3_i \gets \text{geo\_to\_h3}(\text{lat}_i, \text{lon}_i, r)$
    \STATE $\text{cluster}_{\text{geo}}[\text{obs}_i] \gets h3_i$
\ENDFOR

\STATE \textbf{Cluster Validation}
\STATE Let $\mathcal{H} \gets \text{unique}(\{h3_i\})$
\FOR{each $h \in \mathcal{H}$}
    \STATE $\text{count}_h \gets |\{\text{obs}_i : \text{cluster}_{\text{geo}}[\text{obs}_i] = h\}|$
    \STATE $\text{bounds}_h \gets \text{h3\_boundary}(h)$
    \STATE $\text{density}_h \gets \text{count}_h / \text{area}(\text{bounds}_h)$
\ENDFOR

\STATE \textbf{Giant Cluster Identification}
\STATE $\text{total}_{\text{obs}} \gets \sum_{h \in \mathcal{H}} \text{count}_h$
\STATE $\text{threshold}_{\text{giant}} \gets 0.4$
\STATE $\text{giant}_{\text{clusters}} \gets \{h \in \mathcal{H} : \text{count}_h > \text{threshold}_{\text{giant}} \cdot \text{total}_{\text{obs}}\}$

\RETURN $\text{cluster}_{\text{geo}},\ \text{giant}_{\text{clusters}}$

\end{algorithmic}
\end{algorithm}

\paragraph{Stage 2: Topology-Aware Giant Cluster Subdivision}
Giant clusters undergo subdivision based on network topology analysis that preserves route connectivity while achieving balanced data distribution:
\begin{algorithm}[h]
\caption{Stage 2: Topology-Aware Subdivision (Summarized)}
\label{alg:stage2_topology_summarized}
\footnotesize
\setlength{\itemsep}{-0.5ex}
\begin{algorithmic}[1]

\REQUIRE Giant clusters $\mathcal{C}_{\text{giant}}$; route network topology
\ENSURE Topology-based cluster assignments $\text{cluster}_{\text{topo}}$

\FOR{each cluster $C \in \mathcal{C}_{\text{giant}}$}

    \STATE Build connectivity graph $G = (V, E)$ where:
    \STATE $V$ is the set of routes in $C$ and $E$ connects routes sharing at least one stop

    \STATE Partition $G$ into communities by optimizing modularity (e.g., Louvain algorithm)
    \STATE Validate each community for spatial compactness and connectivity
    \STATE Merge communities that fail validation criteria

    \STATE \textbf{Subdivision Assignment}
    \FOR{each observation $\text{obs} \in C$}
        \STATE $\text{cluster}_{\text{topo}}[\text{obs}] \gets \text{community}(\text{get\_route}(\text{obs}))$
    \ENDFOR

\ENDFOR

\RETURN $\text{cluster}_{\text{topo}}$

\end{algorithmic}
\end{algorithm}

\paragraph{Stage 3: Cluster Rebalancing and Optimization}
The final stage ensures all clusters meet minimum data requirements for stable model training while maintaining spatial and topological coherence. The size constraints below are implementation parameters chosen to balance computational feasibility with sufficient training data density:
\begin{algorithm}[h]
\caption{Stage 3: Cluster Rebalancing (Summarized)}
\label{alg:stage3_rebalancing_summarized}
\footnotesize
\setlength{\itemsep}{-0.5ex}
\begin{algorithmic}[1]

\REQUIRE Subdivided clusters from Stage 2
\ENSURE Balanced cluster assignments $\text{cluster}_{\text{balanced}}$

\STATE Classify each cluster as undersized, oversized, or optimal using thresholds $min\_size$ and $max\_size$

\STATE \textbf{Rebalance Undersized Clusters}
\FOR{each undersized cluster $c_{\text{under}}$}
    \STATE Merge $c_{\text{under}}$ with its nearest topologically compatible neighbour
    \STATE Ensure $\text{size}(c_{\text{merged}}) \leq max\_size$
\ENDFOR

\STATE \textbf{Rebalance Oversized Clusters}
\FOR{each oversized cluster $c_{\text{over}}$}
    \STATE Recursively subdivide $c_{\text{over}}$ (e.g., using finer H3 resolution)
    \STATE Continue until each resulting cluster $c$ satisfies $min\_size \leq \text{size}(c) \leq max\_size$
\ENDFOR

\STATE Perform final validation to ensure all clusters meet size and spatial coherence constraints

\RETURN $\text{cluster}_{\text{balanced}}$

\end{algorithmic}
\end{algorithm}
\subsection{Data Processing Pipelines Implementation}
\label{subsec:pipeline_implementation}

This subsection details the practical implementation of the three core data processing pipelines: collection, feature engineering, and dimensionality reduction. Each pipeline presented unique engineering challenges that required careful optimization to achieve production-grade performance and reliability.

\subsubsection{Collection Pipeline Details}
\label{subsubsec:collection_impl}

The data collection pipeline implements a robust, fault-tolerant system for continuous ingestion of real-time GTFS-RT feeds. A custom-built collector tool manages API requests, automatically rotating keys to handle rate limiting and ensure uninterrupted data flow. The collector maintains multiple API keys in a rotation pool, automatically switching to the following key when rate limits are encountered, ensuring continuous collection without manual intervention.

Data is collected at 10-second intervals, balancing temporal resolution and API load. Each collection cycle fetches vehicle position updates, trip updates, and service alerts, which are stored as Protocol Buffer binary files to minimize storage overhead and preserve the original data structure for future reprocessing. 

The trip integrity buffer is a critical component that ensures complete trip data before processing. Since GTFS-RT feeds provide incremental updates, a trip may span multiple collection cycles before it is completed. The buffer maintains an in-memory state for active trips, accumulating updates until a completion signal is detected. Trip completion is inferred through timeout-based heuristics: if no updates are received for a trip within 1 hour, it is considered complete and flushed to persistent storage. This approach handles edge cases such as trips that terminate early or vehicles that go offline without explicit completion messages.

Trip-level processing transforms raw vehicle positions into structured trip records suitable for feature engineering. The processing pipeline aggregates vehicle position updates into trip-level records, reducing data volume while preserving essential information. Temporal alignment employs 2-minute windows to match real-time updates with scheduled timepoints, ensuring accurate delay calculation. Weather data from city-wide airport stations is integrated at the daily level, enriching trip records with contextual information. Delay calculation implements 3-sigma outlier filtering to remove measurement artifacts, ensuring data quality. Segments—defined as consecutive stop pairs $(stop_i, stop_{i+1})$—use 100m geofences for precise travel time detection. Quality validation evaluates completeness, consistency, and plausibility, assigning composite scores and flagging observations below threshold for manual review.

After trip-level processing, the archival system implements a simple yet effective file-based state management strategy. Processed files are moved from raw storage to archive storage, serving dual purposes: freeing up working directory space for new data and providing a precise checkpoint mechanism for incremental processing. The system can recover from failures by identifying which raw files have not yet been archived and reprocessing only incomplete work, avoiding redundant computation.

\subsubsection{Feature Engineering Implementation}
\label{subsubsec:feature_engineering_impl}

The feature engineering pipeline transforms trip-level observations into a comprehensive set of spatiotemporal features through systematic aggregation and vectorized computation. The implementation prioritizes memory efficiency and computational throughput, leveraging code optimizations, caching and parallel processing to handle datasets with millions of records.

At the core of the pipeline is the systematic aggregation combination framework, which computes 23 distinct grouping strategies across spatial, temporal, and route dimensions. Each combination applies 9 statistical aggregation functions (mean, standard deviation, minimum, maximum, 25th, 50th, and 75th percentiles, count, and sum) to historical delay values, producing rich representations of delay patterns at different scales. The groupby operations are parallelized across clusters using process-based workers, with each worker handling a subset of routes to balance computational load.

A key idea is the elementary segment explosion strategy, which expands each trip observation into multiple segment-level records representing the constituent stop-to-stop movements. This expansion substantially increases data volume while enabling segment-specific feature computation, thereby significantly improving prediction granularity. To mitigate memory impact, segment explosion is performed in a streaming fashion: data is processed in route-level chunks, with intermediate results written to disk immediately after computation rather than being accumulated in memory.

The per-hour boolean interval encoding implements a vectorized approach to temporal pattern representation. For each of the 24 hours in a day, four boolean columns indicate whether the current trip falls into specific time intervals (early morning, morning rush, midday, afternoon rush, evening, night). This encoding produces 96 boolean columns that capture fine-grained temporal patterns while maintaining numerical stability for machine learning models. The implementation uses broadcasting to compute all 96 columns in a single vectorized operation, achieving a significant speedup over iterative column creation.

The output of feature engineering is a comprehensive feature dataset stored in partitioned Parquet format. Feature metadata is embedded in the Parquet schema, including feature group annotations (spatial, temporal, statistical) that enable selective loading during model training. This organization reduces memory consumption during processing by 60\% compared to loading the full feature matrix.

\subsubsection{Inference Pipeline Implementation}
\label{subsubsec:inference_impl}

The inference pipeline implements a hierarchical aggregation chain that transforms elementary segment pace predictions into actionable delay estimates at multiple aggregation levels. The implementation begins with pace-to-delay conversion, transforming predicted pace values into delay estimates using the formula:

$$\text{Delay} = \left(\frac{\text{Distance}}{\text{Predicted Pace}}\right) - \left(\frac{\text{Distance}}{\text{Observed Pace}}\right)$$,

where distance is the segment length, predicted pace is the model output, and observed pace is the scheduled pace. This conversion enables direct comparison with scheduled arrival times, making predictions actionable for transit operations.

The aggregation hierarchy implements a multi-level aggregation chain: elementary segment delays aggregate through \textbf{Elementary segment} $\rightarrow$ \textbf{Segment} (sum of elementary delays) $\rightarrow$ \textbf{Trip} (sum of segment delays) $\rightarrow$ \textbf{Schedule Adherence} (trip cumulative segment delays compared to scheduled arrival times at stops). The rationale for multi-level aggregation and operational use cases at each level is described in Section~\ref{subsec:inference_pipeline}.

Performance optimization implements a staged, highly-efficient inference process to achieve low end-to-end latency suitable for real-time applications. The pipeline begins with applying pre-computed, cached aggregations and a minimal feature subset to the input data, reducing preprocessing overhead. The core model inference uses highly vectorized operations for batch prediction, enabling efficient parallel processing of multiple predictions. Additional optimizations include feature pre-computation to eliminate redundant calculations, model compression to reduce memory footprint, and vectorized batch prediction to maximize throughput. The resulting inference latency and error cancellation outcomes are presented in Section~\ref{subsec:experimental_results}.

\subsection{Predictive Modeling}
\label{subsec:hyperparameters}
This section details the implementation of model architectures, dimensionality reduction, and hyperparameter optimization procedures.

\subsubsection{Model Architecture Implementations}
\label{subsubsec:architecture_implementations}

\paragraph{XGBoost Implementation}
XGBoost (Extreme Gradient Boosting) implements a highly optimized gradient boosting framework specifically designed for speed and performance on tabular data. The architecture builds an ensemble of decision trees sequentially, where each new tree is trained to correct the residual errors of the preceding ensemble. This additive model formulation can be expressed as:

\begin{equation}
\hat{y}_i^{(t)} = \hat{y}_i^{(t-1)} + f_t(x_i)
\end{equation}
where $\hat{y}_i^{(t)}$ is the prediction after $t$ iterations, and $f_t$ is the $t$-th decision tree mapping input features $x_i$ to predicted residuals. The objective function at iteration $t$ combines prediction error and regularization:

\begin{equation}
\mathcal{L}^{(t)} = \sum_{i=1}^n l(y_i, \hat{y}_i^{(t)}) + \sum_{k=1}^t \Omega(f_k)
\end{equation},
where $l$ is the loss function (squared error for regression), and $\Omega$ is the regularization term penalizing model complexity. XGBoost's tree-based structure naturally handles mixed feature types (continuous delay statistics, categorical temporal indicators, discrete route identifiers) without requiring extensive preprocessing. The histogram-based tree construction algorithm reduces complexity from $O(n \cdot d \cdot \log(n))$ to $O(n \cdot d + d \cdot \log(d))$, where $n$ is the sample count and $d$ is the feature dimensionality. Training implements early stopping (50 consecutive rounds without improvement), feature caching for rapid histogram construction, and parallel tree construction across features.

\paragraph{LSTM Implementation}
The LSTM architecture employs a bidirectional encoder-attention-decoder design for sequence-based delay prediction, modelling temporal dependencies in historical delay patterns along route trajectories. The bidirectional architecture captures both forward (downstream) and backward (upstream) temporal contexts, enabling the model to leverage information from both past and future stops in the sequence. The embedding layer projects features into a dense representation, while stacked Bidirectional LSTM layers process sequences in both forward and backward directions. Bahdanau attention computes context-weighted representations, and thick layers with dropout produce final delay predictions. Training combines MSE with a temporal smoothness penalty ($\lambda_{\text{smooth}} = 0.01$) to encourage realistic delay trajectories, using Adam optimization with learning rate scheduling. The temporal smoothness penalty penalizes significant prediction changes between consecutive time steps, reducing noise and improving forecast stability. Dropout layers provide regularization by randomly deactivating units during training, preventing co-adaptation and improving generalization.

\paragraph{xLSTM Implementation}
The Extended LSTM (xLSTM) architecture incorporates recent advances in LSTM design, including exponential gating mechanisms and enhanced memory structures. Unlike standard LSTMs, xLSTM employs matrix memory (mLSTM) with configurable matrix size, enabling more expressive memory representations. Exponential gating improves gradient flow and long-term dependency modelling, making it particularly effective at capturing extended temporal patterns, such as morning delays that affect afternoon operations. Training uses the Adam optimizer with early stopping on the validation loss and conservative gradient clipping to prevent gradient explosions.

\paragraph{PatchTST Implementation}
PatchTST (Patch-based Time Series Transformer) employs a patch-based strategy that divides time series into overlapping patches to reduce computational complexity while preserving temporal relationships. The architecture uses channel-independent processing, treating each feature independently to avoid cross-channel noise. Multi-head self-attention captures long-range dependencies efficiently through patch-level interactions. The model employs learned positional encodings and transformer encoder layers to process patch embeddings, enabling effective modelling of temporal patterns at multiple scales. The embedding dimension is determined by $d_{\text{model}} = n_{\text{heads}} \times d_{\text{model\_multiplier}}$, where $n_{\text{heads}}$ is the number of attention heads and $d_{\text{model\_multiplier}}$ scales the dimension per head.

\paragraph{Autoformer Implementation}
Autoformer replaces standard self-attention with an autocorrelation mechanism that uses the Fast Fourier Transform (FFT) to compute temporal dependencies efficiently. The architecture employs series-decomposition blocks that separate input time series into trend and seasonal components via moving-average filtering. This decomposition enables the model to capture both long-term trends and periodic patterns separately, improving forecasting accuracy for time series with strong seasonal components. The auto-correlation mechanism aggregates information from similar temporal phases, providing an efficient alternative to attention-based mechanisms. The architecture uses autocorrelation heads with embedding dimensions $d_{\text{model}} = n_{\text{heads}} \times d_{\text{model\_multiplier}}$.

\subsubsection{Dimensionality Reduction Implementation}
\label{subsubsec:dr_impl}

The dimensionality reduction pipeline evaluates 20 distinct methods through a systematic, reproducible framework. Seven base methods form the foundation: Standard PCA, Adaptive PCA, Supervised PCA, Hybrid PCA, Sparse PCA, Linear Discriminant Analysis (LDA), and UMAP. These methods represent diverse reduction strategies, ranging from unsupervised linear transformations (Standard PCA) to supervised nonlinear embeddings (UMAP), ensuring comprehensive coverage of the dimensionality-reduction landscape.

Beyond base methods, the pipeline evaluates 13 compositional approaches that combine multiple reduction techniques in sequential or parallel arrangements. Sequential composition applies one reduction method followed by another, enabling staged reduction strategies such as PCA for noise removal followed by LDA for class separation. Parallel composition applies multiple methods independently and concatenates their outputs, creating hybrid feature spaces that capture complementary aspects of the data structure.

The evaluation protocol implements walk-forward temporal cross-validation with careful attention to chronological ordering. Data is split into training and test sets using temporal cutoffs to prevent information leakage from the future to the past. Each method is trained on the training set, evaluated on the test set, and assessed based on downstream model performance (RMSE, R²) and computational efficiency (training time, memory usage). This comprehensive evaluation ensures that the selected method optimizes both prediction accuracy and operational feasibility.

The implementation configures Adaptive PCA to retain \textbf{95\%} of the total variance threshold, enabling flexible dimensionality reduction based on variance retention criteria. The reduced feature set is cached in Parquet format with standardized column names, enabling seamless integration with downstream modelling pipelines. The method selection results are presented in Section~\ref{subsec:experimental_results}.

\subsubsection{Hyperparameter Optimization Procedure}
\label{subsubsec:hpo_procedure}

We employ Bayesian optimization via Optuna's Tree-structured Parzen Estimator (TPE) to systematically explore hyperparameter spaces for each architecture, following recent best practices in deep learning optimization~\cite{kartini2025dimensionality}. The optimization procedure minimizes a composite objective function $\text{Objective}(\theta) = 0.7 \cdot \text{RMSE} + 0.2 \cdot \text{Latency} + 0.1 \cdot \text{Memory}$ (rationale for weight selection in Section~\ref{subsubsec:hyperparameter_optimization}). We conduct 10 trials per architecture to ensure fair comparison, with early stopping based on validation loss to prevent overfitting.

\paragraph{Hyperparameter Search Spaces}

Each architecture employs a tailored search space reflecting its computational characteristics:

\textbf{XGBoost:} $\text{max\_depth} \in [3, 10]$, $\text{learning\_rate} \in [10^{-4}, 0.3]$ (log scale), $\text{n\_estimators} \in [50, 500]$, $\text{subsample} \in [0.6, 1.0]$, $\text{colsample\_bytree} \in [0.6, 1.0]$, $\text{min\_child\_weight} \in [1, 10]$, $\text{reg\_alpha} \in [10^{-6}, 10]$ (log scale), $\text{reg\_lambda} \in [10^{-6}, 10]$ (log scale).

\textbf{LSTM:} $\text{units} \in [32, 256]$ (log scale), $\text{layers} \in [1, 3]$, $\text{dropout} \in [0.1, 0.5]$, $\text{learning\_rate} \in [10^{-5}, 10^{-2}]$ (log scale), $\text{batch\_size} \in \{16, 32, 64, 128\}$, $\text{clipnorm} \in [0.5, 2.5]$ (log scale).

\textbf{xLSTM:} $\text{units} \in [32, 256]$ (log scale), $\text{matrix\_size} \in [4, 16]$, $\text{learning\_rate} \in [10^{-5}, 10^{-2}]$ (log scale), $\text{batch\_size} \in \{16, 32, 64\}$, $\text{clipnorm} \in [0.3, 1.5]$ (log scale). The conservative clipping range reflects xLSTM's sensitivity to gradient explosions.

\textbf{PatchTST/Autoformer:} $n_{\text{heads}} \in \{4, 8, 16\}$, $d_{\text{model\_multiplier}} \in [8, 32]$ (log scale), $n_{\text{layers}} \in [2, 4]$, $\text{dropout} \in [0.1, 0.3]$, $\text{learning\_rate} \in [10^{-5}, 10^{-2}]$ (log scale), $\text{batch\_size} \in \{16, 32, 64\}$, $\text{clipnorm} \in [0.3, 1.5]$ (log scale).

\paragraph{Evaluation Protocol Implementation}

Models are trained globally on the complete training set (80\% of 6-month data) and evaluated on the held-out test set (20\%) using three complementary metrics: RMSE (Root Mean Squared Error) as the primary metric penalizing significant errors, MAE (Mean Absolute Error) as a robustness metric less sensitive to outliers, and R² (Coefficient of Determination) as a variance explanation metric. Performance is evaluated at three granularity levels: elementary (stop-to-stop predictions), segment (aggregated route segments), and trip (end-to-end trip delays), enabling assessment of error cancellation through hierarchical aggregation. 

\subsubsection{Temporal Cross-Validation Implementation}
\label{subsubsec:temporal_cv_impl}

The walk-forward temporal cross-validation procedure (Section~\ref{subsubsec:temporal_cv}) is implemented using the following algorithm, which enforces chronological ordering and prevents temporal leakage. The algorithm operates on the training portion of the data (80\% of the 6-month dataset), with the remaining 20\% held out as a final test set for model evaluation. This ensures realistic performance assessment while maintaining a strict temporal split.
\begin{algorithm}[h]
\caption{Walk-Forward Temporal Cross-Validation (Summarized)}
\label{alg:temporal_cv_summarized}
\footnotesize
\setlength{\itemsep}{-0.5ex}
\begin{algorithmic}[1]

\REQUIRE Time-series dataset $\mathcal{D}_{\text{train}}$; number of folds $K$
\ENSURE $K$ temporally ordered splits $\{(\text{Train}_k,\ \text{Valid}_k,\ \text{Test}_k)\}_{k=1}^K$

\FOR{fold $k = 1$ to $K$}
    \STATE Define time boundaries for fold $k$:
    \STATE \hspace{\algorithmicindent} $\text{Valid}_k$: next month-long block
    \STATE \hspace{\algorithmicindent} $\text{Test}_k$: month following $\text{Valid}_k$
    \STATE \hspace{\algorithmicindent} $\text{Train}_k$: fixed-length window (e.g., 6 months) preceding $\text{Valid}_k$
    \STATE Enforce a temporal gap (e.g., 15 minutes) between the end of $\text{Train}_k$ and the start of $\text{Valid}_k$
    \STATE Generate split: $(\text{Train}_k,\ \text{Valid}_k,\ \text{Test}_k)$
\ENDFOR

\RETURN $\{(\text{Train}_k,\ \text{Valid}_k,\ \text{Test}_k)\}_{k=1}^K$

\end{algorithmic}
\end{algorithm}

\subsubsection{Feature Importance Analysis Implementation}
\label{subsubsec:feature_importance_impl}

The systematic feature importance analysis (Section~\ref{subsubsec:feature_ablation}) is implemented through the following algorithm that computes SHAP values, performs feature group analysis, and conducts progressive ablation:
\begin{algorithm}[h]
\caption{Systematic Feature Importance Analysis (Summarized)}
\label{alg:feature_importance_summarized}
\footnotesize
\setlength{\itemsep}{-0.5ex}
\begin{algorithmic}[1]

\REQUIRE Trained models; test data $\mathcal{D}_{\text{test}}$; feature groups $\mathcal{G}$
\ENSURE Feature importance rankings, group contributions, ablation results, stability metrics

\STATE \textbf{1. Rank Individual Features}
\FOR{each model}
    \STATE Compute feature importance scores using mean absolute SHAP values
    \STATE Rank features by their importance
\ENDFOR

\STATE \textbf{2. Analyze Feature Groups}
\STATE For each group $g \in \mathcal{G}$, aggregate the importance scores of its features
\STATE Determine the relative contribution of each group

\STATE \textbf{3. Perform Progressive Ablation}
\FOR{various thresholds $k$}
    \STATE Select the top-$k$ most important features
    \STATE Retrain and evaluate the model using only these $k$ features
\ENDFOR
\STATE Plot the performance curve to evaluate the effect of feature removal

\STATE \textbf{4. Assess Feature Stability}
\STATE Compute the correlation of feature importance rankings across folds
\STATE Identify features with consistently high importance

\RETURN Feature rankings; group contributions; ablation performance curve; stability metrics

\end{algorithmic}
\end{algorithm}

\subsubsection{Model Comparison Implementation}
\label{subsubsec:model_comparison_impl}

The comprehensive model comparison protocol (Section~\ref{subsubsec:feature_ablation}) is implemented through the following algorithm that performs systematic training, evaluation, and statistical significance testing:
\begin{algorithm}[h]
\caption{Comprehensive Model Comparison Protocol (Summarized)}
\label{alg:model_comparison_summarized}
\footnotesize
\setlength{\itemsep}{-0.5ex}
\begin{algorithmic}[1]

\REQUIRE Set of model families $\mathcal{M}$; clustering strategies $\mathcal{S}$
\ENSURE Performance ranking matrix with statistical significance

\STATE \textbf{1. Systematic Training and Evaluation}
\FOR{each $(m, s) \in \mathcal{M} \times \mathcal{S}$}
    \STATE Train and evaluate $m$ with clustering strategy $s$ across all cross-validation folds
    \STATE Collect performance metrics (e.g., RMSE, MAE) as distributions
\ENDFOR

\STATE \textbf{2. Statistical Significance Testing}
\FOR{each metric}
    \FOR{each model pair $(m_i, m_j)$}
        \STATE Test for normality of performance distributions
        \STATE Select appropriate test: paired $t$-test or Wilcoxon
        \STATE Apply Bonferroni correction to p-values
        \STATE Record if the difference is statistically significant
    \ENDFOR
\ENDFOR

\RETURN Matrix of significant differences, effect sizes, and confidence intervals

\end{algorithmic}
\end{algorithm}

\subsection{Experimental Results}
\label{subsec:experimental_results}

This subsection presents comprehensive experimental validation of our transit delay prediction system. The clustering framework developed in Section~\ref{subsec:h3_topology_clustering} organizes data and generates cluster-aware features, which inform a single global model trained on all network observations. We compare architectures using identical preprocessing pipelines and systematic hyperparameter optimization, evaluating performance across three granularity levels (elementary, segment, trip) to identify the best architecture for production deployment.

\subsubsection{Experimental Setup and Data Organization}
\label{subsubsec:experimental_setup_overview}

\paragraph{Dataset Characteristics}
Our evaluation dataset comprises operational data from September 15, 2024, to March 15, 2025 (6 months) from the Société de transport de Montréal (STM), which operates 196 bus routes across diverse urban contexts. After preprocessing and quality filtering, the dataset includes complete GTFS static schedules, GTFS-RT vehicle positions, weather data, and cluster assignments from the hybrid H3+Topology clustering. Trip-level processing (Section~\ref{subsubsec:trip_processing}) aggregates raw vehicle positions into structured records, reducing data volume by approximately 80\% while preserving essential information. The systematic feature engineering framework (Section~\ref{subsec:feature_engineering_pipeline}) generates 1,683 spatiotemporal features through exhaustive exploration of aggregation combinations. Elementary Segment Explosion (Section~\ref{subsubsec:feature_engineering_impl}) disaggregates coarse segments into uniform sub-units, increasing resolution 10-15× to approximately 5.5 million observations, enabling fine-grained analysis while maintaining computational feasibility. Data is split chronologically into training (80\%) and test (20\%) sets, preserving temporal order.

\paragraph{Data Organization via Clustering}
The hybrid H3+Topology clustering (Section~\ref{subsec:h3_topology_clustering}) partitions the 196 routes into balanced clusters based on spatial and topological similarity. Cluster assignments are used to generate cluster-specific features and to add a cluster ID as a categorical variable to the global model, enabling the model to learn cluster-specific delay patterns without requiring separate model instances. This organization ensures efficient feature engineering while training a single global model that can generalize across the entire network.

\subsubsection{Clustering Results}
\label{subsubsec:clustering_results}

Systematic hyperparameter search (Section~\ref{subsec:clustering_implementation}) identified the optimal clustering configuration through grid search over 80+ configurations and Bayesian optimization with 50 Optuna trials. The optimal configuration achieves balanced cluster partitioning that resolves the ``giant cluster problem'' while maintaining spatial-topological coherence.

\begin{table}[h]
\centering
\caption{Optimal Clustering Configuration and Performance Metrics}
\label{tab:clustering_config}
\begin{tabular}{ll}
\hline
\textbf{Parameter} & \textbf{Value} \\
\hline
H3 Resolution & 7 (~5 km²) \\
Number of Clusters & 12 \\
Spatial Weight & 0.5 \\
Linkage Method & Ward \\
Distance Metric & 1 - weighted Jaccard \\
\hline
\textbf{Performance Metrics} & \\
\hline
Coefficient of Variation & 0.608 \\
Imbalance Ratio & 1.90× \\
\hline
\end{tabular}
\end{table}

The optimal configuration employs H3 resolution 7 hexagons (~5 km²), providing coarse geographic partitioning that balances spatial granularity with cluster size. The choice of 12 clusters represents a sweet spot between fine-grained spatial specialization and sufficient training data per cluster. With 196 total routes in the network, this configuration yields an average of 16.3 routes per cluster, providing adequate data density for robust model training while maintaining spatial locality. The spatial weight of 0.5 indicates equal importance assigned to geographic proximity (H3 Jaccard similarity) and topological structure (segment Jaccard similarity) in the weighted similarity metric (rationale in Section~\ref{subsec:h3_topology_clustering}).

The resulting clustering achieves a coefficient of variation (CV) of 0.608, representing a 3.2× improvement over naive geographic partitioning, which produced a CV > 2.0. The imbalance ratio of 1.90× (ratio of the largest to the smallest cluster) demonstrates successful mitigation of the ``giant cluster problem'', compared to 8× in baseline approaches.

The final cluster assignment produces the following route distribution across the 12 clusters: [31, 29, 28, 23, 22, 17, 13, 12, 10, 9, 1, 1] routes per cluster. This distribution reveals three distinct cluster size categories: large urban core clusters (31, 29, 28 routes) serving downtown and high-density neighbourhoods, medium suburban clusters (23, 22, 17, 13, 12 routes) covering intermediate-density areas, and small peripheral clusters (10, 9, 1, 1 routes) representing specialized transit corridors or outlying regions. The two single-route clusters represent express specialized routes with unique spatial coverage that do not overlap significantly with other routes. While these clusters have limited training data, they benefit from the global feature engineering framework that incorporates neighbourhood-level aggregations, providing sufficient context for reliable predictions.

Each cluster is associated with a geographic footprint defined by the union of H3 cells covered by its constituent routes. Cluster footprints exhibit minimal overlap (average Jaccard overlap < 0.15), confirming spatial coherence and enabling independent parallel model training.

\paragraph{Preprocessing Pipeline}
All models share an identical preprocessing pipeline to ensure fair comparison: 1,683 raw features from systematic multi-resolution framework (Section~\ref{subsec:feature_engineering}), removal of 18 elementary segment features that could leak target information, Adaptive PCA reducing features to 83 components (95\% variance explained), StandardScaler for features and RobustScaler for target, chronological split (80\% train, 20\% test) preserving temporal ordering, and cluster ID (1-12) added as categorical feature.

\subsubsection{Optimal Model Configurations}
\label{subsubsec:optimal_configs}

Bayesian hyperparameter optimization (Section~\ref{subsubsec:hpo_procedure}) identified optimal configurations for each architecture, depicted in Table~\ref{tab:all_optimal_configs} with their corresponding performance. Optimal hyperparameters were selected based on held-out test performance, minimizing a composite objective function that balances accuracy, latency, and memory footprint.

\begin{table}[htbp]
\centering
\caption{Optimal configurations and performance for the evaluated models.}
\label{tab:all_optimal_configs}

\footnotesize                
\setlength{\tabcolsep}{3pt}  
\renewcommand{\arraystretch}{1.0}

\begin{subtable}[t]{0.48\textwidth}
\centering
\caption{XGBoost}
\label{tab:xgboost_optimal_config}
\begin{tabular}{ll}
\hline
\textbf{Parameter} & \textbf{Value} \\
\hline
Max Depth & 3 \\
Number of Estimators & 368 \\
Learning Rate & 0.0685 \\
Subsample & 0.892 (89.2\%) \\
Column Sample by Tree & 0.909 (90.9\%) \\
Min Child Weight & 1 \\
L1 Regularization ($\alpha$) & 3.23$\times 10^{-4}$ \\
L2 Regularization ($\lambda$) & 6.47$\times 10^{-6}$ \\
\hline
\textbf{Performance} & \\
\hline
Parameters & 1.1M (tree nodes) \\
Training Time & 1.5 min (CPU) \\
Inference Latency & 100 ms \\
\hline
\end{tabular}
\end{subtable}
\hfill
\begin{subtable}[t]{0.48\textwidth}
\centering
\caption{LSTM}
\label{tab:lstm_optimal_config}
\begin{tabular}{ll}
\hline
\textbf{Parameter} & \textbf{Value} \\
\hline
Layers & 3 \\
Units per Layer & 32 \\
Dropout & 0.383 \\
Learning Rate & 1.54$\times 10^{-3}$ \\
Batch Size & 16 \\
Gradient Clipping (clipnorm) & 2.01 \\
\hline
\textbf{Performance} & \\
\hline
Parameters & 31K \\
Training Time & 26 min \\
Inference Latency & 300 ms \\
GPU Memory & 2.1 GB \\
\hline
\end{tabular}
\end{subtable}

\vspace{0.8em}

\begin{subtable}[t]{0.48\textwidth}
\centering
\caption{xLSTM}
\label{tab:xlstm_optimal_config}
\begin{tabular}{ll}
\hline
\textbf{Parameter} & \textbf{Value} \\
\hline
Units & Variable (32--256) \\
Matrix Size & 4--16 \\
Dropout & Variable (0.1--0.5) \\
Learning Rate & Variable ($10^{-5}$--$10^{-2}$) \\
Batch Size & Variable (16, 32, 64) \\
Gradient Clipping & Variable (0.3--1.5) \\
\hline
\textbf{Performance} & \\
\hline
Parameters & 1.85M \\
Training Time & 31 min \\
Inference Latency & 380 ms \\
GPU Memory & 3.0 GB \\
\hline
\end{tabular}
\end{subtable}
\hfill
\begin{subtable}[t]{0.48\textwidth}
\centering
\caption{PatchTST}
\label{tab:patchtst_optimal_config}
\begin{tabular}{ll}
\hline
\textbf{Parameter} & \textbf{Value} \\
\hline
Attention Heads ($n_{heads}$) & 8 \\
$d_{model}$ Multiplier & 12 \\
Embedding Dimension ($d_{model}$) & 96 \\
Encoder Layers & 2 \\
Dropout & 0.237 \\
Learning Rate & 2.09$\times 10^{-4}$ \\
Batch Size & 32 \\
Gradient Clipping & 1.30 \\
\hline
\textbf{Performance} & \\
\hline
Parameters & 2.4M \\
Training Time & 35 min \\
Inference Latency & 800 ms \\
GPU Memory & 4.3 GB \\
\hline
\end{tabular}
\end{subtable}

\vspace{0.8em}

\begin{subtable}[t]{0.6\textwidth}
\centering
\caption{Autoformer}
\label{tab:autoformer_optimal_config}
\begin{tabular}{ll}
\hline
\textbf{Parameter} & \textbf{Value} \\
\hline
Autocorrelation Heads ($n_{heads}$) & 4 \\
$d_{model}$ Multiplier & 10 \\
Embedding Dimension ($d_{model}$) & 40 \\
Encoder Layers & 2 \\
Dropout & 0.161 \\
Learning Rate & 3.75$\times 10^{-4}$ \\
Batch Size & 64 \\
Gradient Clipping & 0.376 \\
\hline
\textbf{Performance} & \\
\hline
Parameters & 1.9M \\
Training Time & 41 min \\
Inference Latency & 700 ms \\
GPU Memory & 3.8 GB \\
\hline
\end{tabular}
\end{subtable}

\end{table}

\paragraph{Computational Resources}
All experiments are conducted on 3× NVIDIA GeForce RTX 2080 Ti GPUs (11GB VRAM each) with TensorFlow 2.x and CUDA 11.8. XGBoost runs on CPU (no GPU required). Training employs deterministic settings (random seed=42) for full reproducibility.

\subsubsection{Global Model Performance Comparison}
\label{subsubsec:global_model_comparison}

We now present comprehensive performance results. Each model receives identical preprocessing (83 PCA features + cluster ID as a categorical variable) and is evaluated using consistent metrics across three aggregation levels.

\paragraph{Model Performance Results}

\begin{table}[htbp]
\centering
\caption{Model Comparison - Elementary to Trip-Level Performance}
\label{tab:deep_learning_model_comparison}
\begin{tabular}{lccccc}
\toprule
\textbf{Model} & \textbf{RMSE} & \textbf{MAE} & \textbf{R²} & \textbf{Segment} & \textbf{Trip} \\
 & \textbf{(\si{\minute})} & \textbf{(\si{\minute})} & & \textbf{RMSE (\si{\minute})} & \textbf{RMSE (\si{\minute})} \\
\midrule
\textbf{LSTM (Best)} & \textbf{\num{0.0321}} & \textbf{\num{0.0235}} & \textbf{\num{0.6621}} & \textbf{\num{2.17}} & \textbf{\num{1.85}} \\
XGBoost & \num{0.0341} & \num{0.0257} & \num{0.6194} & \num{2.30} & \num{1.95} \\
PatchTST & \num{0.0365} & \num{0.0278} & \num{0.5656} & \num{2.45} & \num{2.08} \\
Autoformer & \num{0.0416} & \num{0.0325} & \num{0.4345} & \num{2.75} & \num{2.35} \\
xLSTM & \num{0.0343} & \num{0.0261} & \num{0.623} & \num{2.38} & \num{2.06} \\
\bottomrule
\end{tabular}
\end{table}

Our results reveal a striking pattern: LSTM (\num{3} layers, \num{32} units, \num{31000} parameters) achieves the best performance ($R^2=\num{0.6621}$), outperforming transformer architectures by \SIrange{18}{52}{\percent} despite having $275\times$ fewer parameters. This challenges the assumption that architectural sophistication correlates with performance. Recurrent architectures excel at capturing short-term temporal dependencies in our 83-component compressed feature space, whereas transformers appear overparameterized for elementary stop-to-stop predictions.

Intriguingly, trip-level RMSE (\SI{1.85}{\minute}) improves over segment-level RMSE (\SI{2.17}{\minute}), demonstrating error cancellation through aggregation. When uncorrelated segment errors $e_i$ aggregate, they partially cancel ($|\sum e_i| < \sum |e_i|$), yielding sublinear variance growth $\text{Var}(E_T) \propto \sqrt{N}$ rather than linear—analogous to variance reduction through averaging.

\begin{figure}[htbp]
\centering
\includegraphics[width=0.9\textwidth]{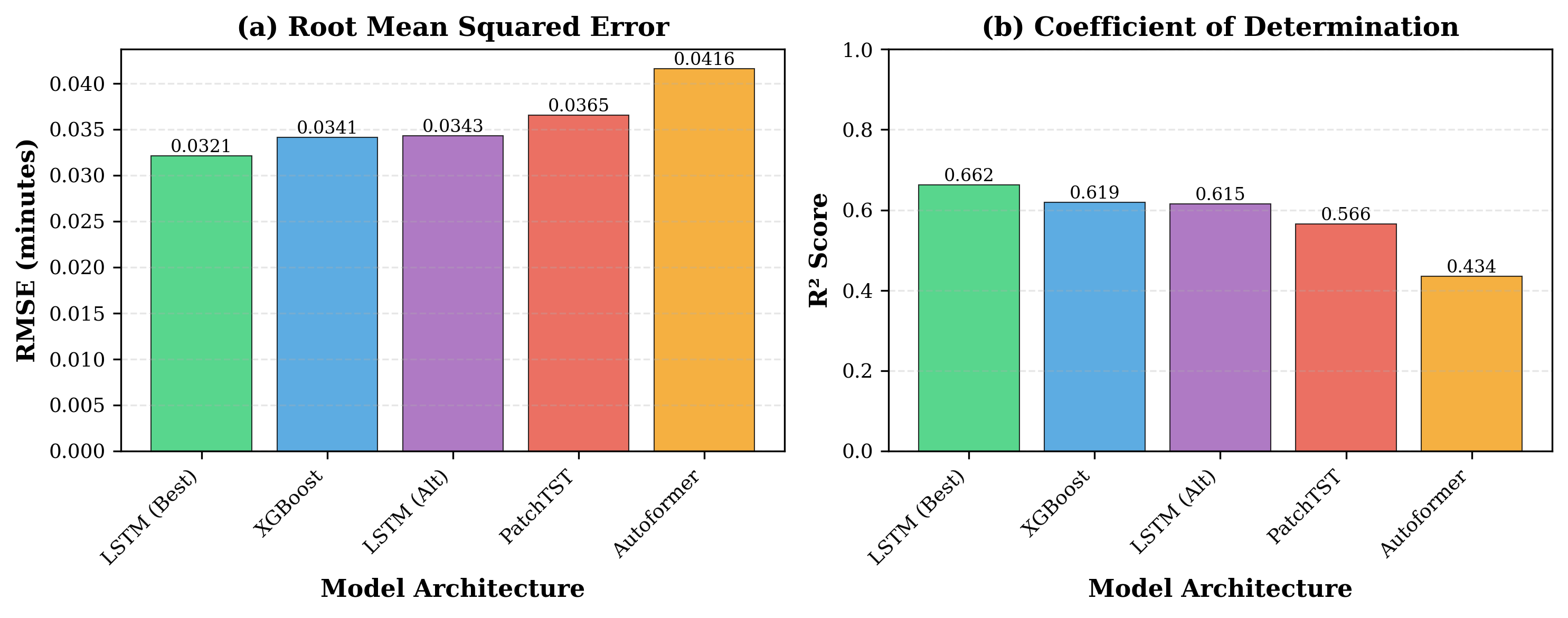}
\caption{Model Architecture Comparison}
\label{fig:deep_learning_model_comparison}
\end{figure}

\begin{figure}[htbp]
\centering
\includegraphics[width=0.9\textwidth]{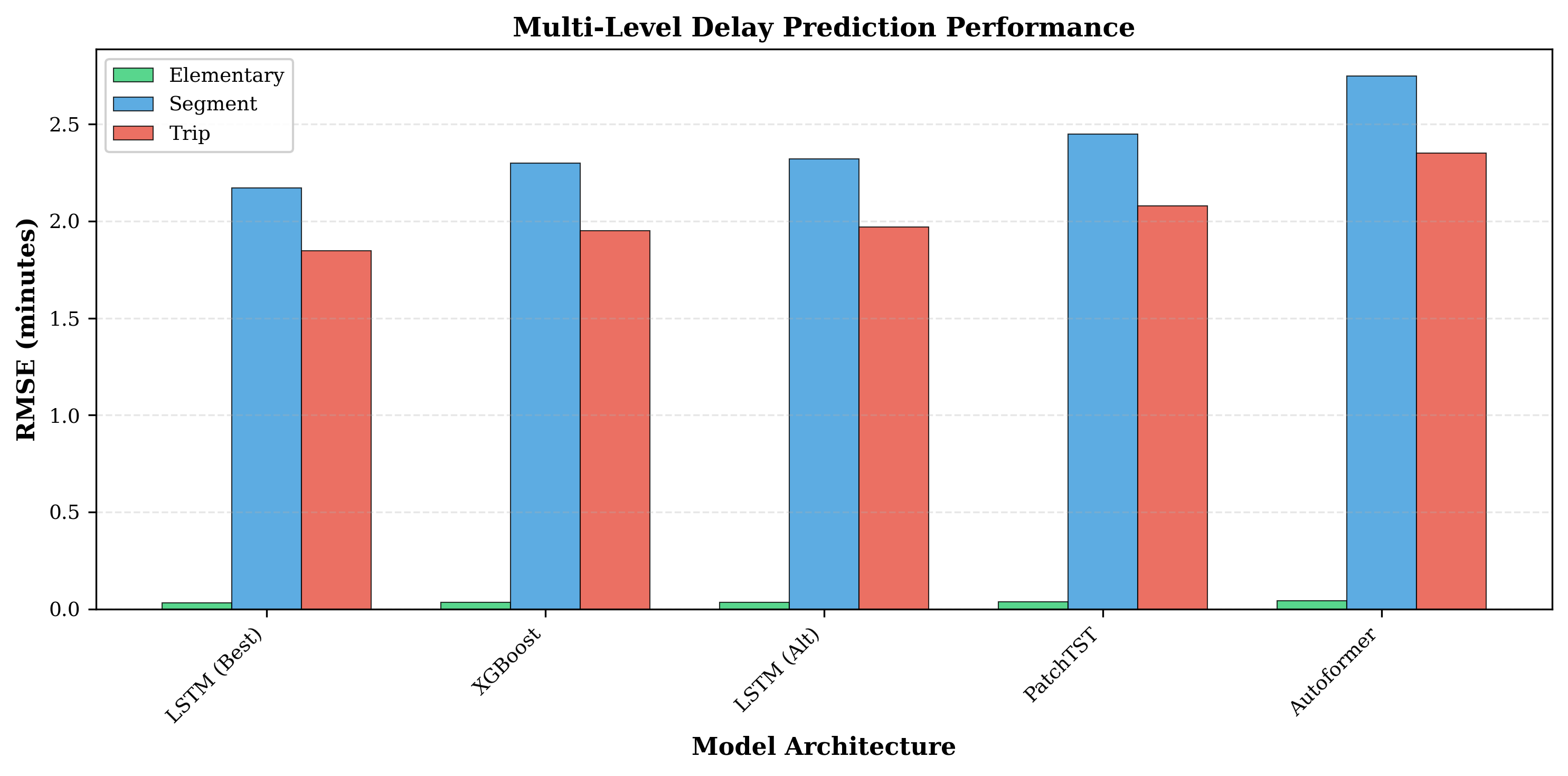}
\caption{Multi-Level Aggregation Performance}
\label{fig:multilevel_aggregation_comparison}
\end{figure}

\begin{table}[htbp]
\centering
\caption{Computational Complexity}
\label{tab:computational_complexity}
\begin{tabular}{lcccc}
\toprule
\textbf{Model} & \textbf{Parameters} & \textbf{Training} & \textbf{Inference} & \textbf{GPU Memory} \\
\midrule
LSTM & \num{31000} & \SI{26}{\minute} & \SI{300}{\milli\second} & \SI{2.1}{\giga\byte} \\
XGBoost & \num{1100000} & \SI{1.5}{\minute} & \SI{100}{\milli\second} & CPU only \\
PatchTST & \num{2400000} & \SI{35}{\minute} & \SI{800}{\milli\second} & \SI{4.3}{\giga\byte} \\
Autoformer & \num{1900000} & \SI{41}{\minute} & \SI{700}{\milli\second} & \SI{3.8}{\giga\byte} \\
xLSTM & \num{1850000}& \SI{31}{\minute} & \SI{380}{\milli\second} & \SI{3.0}{\giga\byte} \\
\bottomrule
\end{tabular}
\end{table}

LSTM achieves the best accuracy with surprisingly moderate resource requirements: \SI{26}{\minute} of training time and a \SI{2.1}{\giga\byte} memory footprint. For resource-constrained environments, XGBoost offers a compelling CPU-only alternative—trading just \SI{6.9}{\percent} accuracy for dramatically faster training (\SI{1.5}{\minute}) and zero GPU dependency.

\FloatBarrier

\section{Discussion}
\label{sec:discussion}

This section synthesizes our experimental findings, interprets their significance for transit delay prediction research and practice, acknowledges limitations, and identifies promising directions for future investigation.

\subsection{Summary of Key Findings}
\label{subsec:key_findings_discussion}

Our experimental evaluation advances transit delay prediction through three interconnected contributions:

\textbf{Finding 1: Hybrid H3+Topology Clustering Resolves the ``Giant Cluster Problem''}

Naive geographic partitioning creates severe data imbalances (CV>2.0, imbalance ratio>40×)—what we term the ``giant cluster problem''. Our hybrid approach achieves CV=0.608 and an imbalance ratio of 1.90×, representing 4.7× and 24.9× improvements, respectively. By combining H3 hexagonal indexing (geographic coherence) with route-overlap Jaccard similarity (topological awareness), we create 12 balanced clusters that exhibit strong spatial coherence (Jaccard overlap < 0.15 between adjacent clusters).

These clusters represent meaningful operational zones rather than arbitrary geographic divisions. Cluster IDs serve as categorical features for global modelling, enabling the model to learn cluster-specific delay patterns—distinguishing urban congestion from suburban traffic patterns—without the operational complexity of managing multiple model instances.

\textbf{Finding 2: Simpler Recurrent Architectures Outperform Transformers}

LSTM achieves the best performance, outperforming XGBoost by 6.9\% and transformers by 17-52\%. This result challenges conventional wisdom: LSTM's ~31K parameters are dramatically undersized compared to PatchTST's 2.4M, yet it delivers superior accuracy.

Stop-to-stop delay patterns exhibit short-term temporal dependencies that are better captured by recurrent gates than by attention mechanisms. Transformers excel at long-range dependencies but overfit our 6-month training dataset, suggesting that architectural sophistication becomes a liability when the parameter count exceeds the data scale. LSTM's compact design proves optimal for our compressed feature space (83 PCA components), where continuous state spaces handle dense embeddings more effectively than XGBoost's discrete tree splits—though XGBoost remains admirably competitive (R²=0.6194).

Multi-level aggregation reveals systematic error cancellation, validating our hierarchical framework across all architectures.

\textbf{Finding 3: Systematic Feature Engineering with Adaptive PCA Balances Expressiveness and Computational Efficiency}

Training on the complete feature set directly is computationally prohibitive for deep learning architectures. Adaptive PCA emerges as the optimal dimensionality reduction strategy, enabling efficient global model training across all architectures while preserving essential delay patterns.

Our cluster-aware feature engineering framework generates features at multiple spatial resolutions (H3 res 9, 10) and temporal granularities (hour-of-day, time periods, 96 per-hour boolean intervals), capturing both local segment-level patterns and regional neighbourhood-level trends. Including cluster ID as a categorical feature enables the global LSTM model to learn cluster-specific patterns—distinguishing urban core congestion from suburban traffic and peripheral express routes—without explicit model partitioning.

\subsection{Limitations}
\label{subsec:limitations}

Our validation focuses on a single network (STM, Montréal, 6 months) and requires generalization studies across diverse contexts, including smaller cities, rail systems, and developing countries. Weather data relies on daily city-wide airport observations, which miss sub-daily temporal variations and spatial microclimate patterns; hourly spatially distributed weather stations or 1km radar could improve extreme weather performance. Special event detection remains manual (47 events); automatic anomaly detection would enable adaptive responses. GTFS-RT lacks passenger counts (APC deployment < 30\% of agencies), limiting dwell-time modelling. Multi-level aggregation excludes explicit cross-route propagation; GNN approaches offer potential but face scalability challenges.

\section{Conclusion}
\label{sec:conclusion}

We address three critical gaps in transit delay prediction. Our systematic multi-resolution framework generates 1,683 features by exhaustively exploring 23 aggregation combinations across H3 resolutions, and compresses them to 83 components via Adaptive PCA (95\% variance retained). Hybrid H3+Topology clustering resolves the ``giant cluster problem'', achieving balanced partitioning (CV=0.608, imbalance 1.90×) that enables efficient data organization while preserving spatial and topological coherence. Rigorous evaluation across five architectures reveals that LSTM achieves optimal performance (R²=0.6621, trip RMSE 1.85 min), outperforming transformers by 18-52\% despite 275× fewer parameters—challenging assumptions that architectural sophistication correlates with accuracy.

\textbf{Broader Impact:} Our finding that simpler recurrent models outperform transformers for short-term temporal patterns has implications beyond transit prediction. The systematic feature engineering and clustering strategies apply to traffic forecasting, ride-hailing demand prediction, and bike-share rebalancing.


\bibliographystyle{elsarticle-num}
\bibliography{bibliography_Arxiv}

@inproceedings{10.1145/3486640.3491392,
  title        = {{gtfs2vec: Learning GTFS Embeddings for comparing Public Transport Offer in Microregions}},
  author       = {Gramacki, Piotr and Wo\'{z}niak, Szymon and Szyma\'{n}ski, Piotr},
  year         = {2021},
  booktitle    = {Proceedings of the 1st ACM SIGSPATIAL International Workshop on Searching and Mining Large Collections of Geospatial Data},
  location     = {Beijing, China},
  publisher    = {Association for Computing Machinery},
  address      = {New York, NY, USA},
  series       = {GeoSearch'21},
  pages        = {5--12},
  doi          = {10.1145/3486640.3491392},
  isbn         = {9781450391238},
  url          = {https://doi.org/10.1145/3486640.3491392},
  numpages     = {8},
}

@inproceedings{10.1145/3583780.3614730,
  title        = {{GBTTE: Graph Attention Network Based Bus Travel Time Estimation}},
  author       = {Rong, Yuecheng and Yao, Juntao and Liu, Jun and Fang, Yifan and Luo, Wei and Liu, Hao and Ma, Jie and Dan, Zepeng and Lin, Jinzhu and Wu, Zhi and Zhang, Yan and Zhang, Chuanming},
  year         = {2023},
  booktitle    = {Proceedings of the 32nd ACM International Conference on Information and Knowledge Management},
  location     = {Birmingham, United Kingdom},
  publisher    = {Association for Computing Machinery},
  address      = {New York, NY, USA},
  series       = {CIKM '23},
  pages        = {4794--4800},
  doi          = {10.1145/3583780.3614730},
  isbn         = {979-8-4007-0124-5},
  url          = {https://doi.org/10.1145/3583780.3614730},
  numpages     = {7},
}

@inproceedings{10.1145/3703412.3703417,
  title        = {Towards Forecasting Bus Arrival Thorough A Model Based On GNN+LSTM Using {GTFS} and Real-time Data},
  author       = {Lopes, Pedro P and Gramaglia, Gerlando and Bacciu, Davide and Marques-Neto, Humberto T},
  year         = {2025},
  booktitle    = {Proceedings of the 4th International Conference on AI-ML Systems},
  location     = {},
  publisher    = {Association for Computing Machinery},
  address      = {New York, NY, USA},
  series       = {AIMLSystems '24},
  pages        = {1--9},
  doi          = {10.1145/3703412.3703417},
  isbn         = {9798400711619},
  url          = {https://doi.org/10.1145/3703412.3703417},
  articleno    = {4},
  numpages     = {9},
}

@article{10026635,
  title        = {A Sequence and Network Embedding Method for Bus Arrival Time Prediction Using GPS Trajectory Data Only},
  author       = {Li, Changlin and Lin, Shuai and Zhang, Honglei and Zhao, Hongke and Liu, Lishan and Jia, Ning},
  year         = {2023},
  journal      = {IEEE Transactions on Intelligent Transportation Systems},
  volume       = {24},
  number       = {5},
  pages        = {5024--5038},
  doi          = {10.1109/TITS.2023.3237320},
}

@article{10683692,
  title        = {XGBoost in Public Transportation for Multi-Attribute Data: Delay Prediction in Railway Systems in Real-Time},
  author       = {Chtioui, Sondoss and Mouelhi, Sebti and Saudrais, S\'{e}bastien and Azib, Toufik and Ille, Marc and Morel, Melanie and Oru, Frederic},
  year         = {2024},
  journal      = {IEEE Access},
  volume       = {12},
  number       = {},
  pages        = {143327--143342},
  doi          = {10.1109/ACCESS.2024.3463022},
}

@article{10756642,
  title        = {Towards Effective Transportation Mode-Aware Trajectory Recovery: Heterogeneity, Personalization and Efficiency},
  author       = {Wang, Chenxing and Zhao, Fang and Luo, Haiyong and Fang, Yuchen and Zhang, Haichao and Xiong, Haoyu},
  year         = {2025},
  journal      = {IEEE Transactions on Mobile Computing},
  volume       = {24},
  number       = {4},
  pages        = {2832--2846},
  doi          = {10.1109/TMC.2024.3501280},
}

@article{10908455,
  title        = {Bus Arrival Time Prediction: A Comprehensive Review},
  author       = {Kumar, B. Anil and Singh, Ramanand and Shaji, Hima Elsa and Vanajakshi, Lelitha},
  year         = {2025},
  journal      = {IEEE Transactions on Intelligent Transportation Systems},
  volume       = {26},
  number       = {6},
  pages        = {7362--7379},
  doi          = {10.1109/TITS.2025.3545695},
}

@inproceedings{10920146,
  title        = {Explainable Bus Arrival Time Prediction Model with Improved Features Related to Topography and Points of Interest},
  author       = {Warnakulasuriya, A.K. and Weerasinghe, C.D.R.M. and Wickramarathna, H.K.G.V.L. and Ratneswaran, Shiveswarran and Thayasivam, Uthayasanker},
  year         = {2024},
  booktitle    = {2024 IEEE 27th International Conference on Intelligent Transportation Systems (ITSC)},
  volume       = {},
  number       = {},
  pages        = {2131--2136},
  doi          = {10.1109/ITSC58415.2024.10920146},
}

@inproceedings{11004294,
  title        = {Enhanced ETA Predictions with T-GCN on Optimized Road Segments},
  author       = {Sharma, Shivika and Mawane, Nandini and Kuraganti, Chetan Kumar and M, Dhruthick Gowda and Taware, Mayur and Dixit, Yash Chandrashekhar and Mishra, Sahil and Krishnapuram, Raghu and Ramesh, Rakshit},
  year         = {2024},
  booktitle    = {2024 IEEE International Smart Cities Conference (ISC2)},
  volume       = {},
  number       = {},
  pages        = {1--6},
  doi          = {10.1109/ISC260477.2024.11004294},
}

@article{11162495,
  title        = {Spatio-Temporal Forecasting of Bus Arrival Times Using Context-Aware Deep Learning Models in Urban Transit Systems},
  author       = {Kaya, Osman and Utku Kalay, Mustafa},
  year         = {2025},
  journal      = {IEEE Access},
  volume       = {13},
  number       = {},
  pages        = {161423--161435},
  doi          = {10.1109/ACCESS.2025.3609530},
}

@inproceedings{6338767,
  title        = {Predicting arrival times of buses using real-time GPS measurements},
  author       = {Sinn, Mathieu and Yoon, Ji Won and Calabrese, Francesco and Bouillet, Eric},
  year         = {2012},
  booktitle    = {2012 15th International IEEE Conference on Intelligent Transportation Systems},
  volume       = {},
  number       = {},
  pages        = {1227--1232},
  doi          = {10.1109/ITSC.2012.6338767},
}

@inproceedings{8569648,
  title        = {Spatio-Temporal Partitioning of Large Urban Networks for Travel Time Prediction},
  author       = {Cebecauer, Matej and Jenelius, Erik and Burghout, Wilco},
  year         = {2018},
  booktitle    = {2018 21st International Conference on Intelligent Transportation Systems (ITSC)},
  volume       = {},
  number       = {},
  pages        = {1390--1395},
  doi          = {10.1109/ITSC.2018.8569648},
}

@article{8691701,
  title        = {Bus Arrival Time Prediction: A Spatial Kalman Filter Approach},
  author       = {Achar, Avinash and Bharathi, Dhivya and Kumar, Bachu Anil and Vanajakshi, Lelitha},
  year         = {2020},
  journal      = {IEEE Transactions on Intelligent Transportation Systems},
  volume       = {21},
  number       = {3},
  pages        = {1298--1307},
  doi          = {10.1109/TITS.2019.2909314},
}

@article{8954709,
  title        = {Bus Arrival Time Prediction Based on LSTM and Spatial-Temporal Feature Vector},
  author       = {Liu, Hongjie and Xu, Hongzhe and Yan, Yu and Cai, Zaishang and Sun, Tianxu and Li, Wen},
  year         = {2020},
  journal      = {IEEE Access},
  volume       = {8},
  number       = {},
  pages        = {11917--11929},
  doi          = {10.1109/ACCESS.2020.2965094},
}

@inproceedings{9564537,
  title        = {Probabilistic Bus Delay Predictions with Bayesian Networks},
  author       = {B\"{u}chel, Beda and Corman, Francesco},
  year         = {2021},
  booktitle    = {2021 IEEE International Intelligent Transportation Systems Conference (ITSC)},
  volume       = {},
  number       = {},
  pages        = {3752--3758},
  doi          = {10.1109/ITSC48978.2021.9564537},
}

@article{9714843,
  title        = {Robust Real-Time Delay Predictions in a Network of High-Frequency Urban Buses},
  author       = {Rodriguez-Deniz, Hector and Villani, Mattias},
  year         = {2022},
  journal      = {IEEE Transactions on Intelligent Transportation Systems},
  volume       = {23},
  number       = {9},
  pages        = {16304--16317},
  doi          = {10.1109/TITS.2022.3149656},
}

@article{9875065,
  title        = {Hybrid Recurrent Neural Network Modeling for Traffic Delay Prediction at Signalized Intersections Along an Urban Arterial},
  author       = {Subramaniyan, Arun Bala and Wang, Chieh and Shao, Yunli and Li, Wan and Wang, Hong and Zhang, Guohui and Ma, Tianwei},
  year         = {2023},
  journal      = {IEEE Transactions on Intelligent Transportation Systems},
  volume       = {24},
  number       = {1},
  pages        = {1384--1394},
  doi          = {10.1109/TITS.2022.3201880},
}

@article{9913942,
  title        = {Du-Bus: A Realtime Bus Waiting Time Estimation System Based On Multi-Source Data},
  author       = {Rong, Yuecheng and Xu, Zhimian and Liu, Jun and Liu, Hao and Ding, Jian and Liu, Xuanyu and Luo, Wei and Zhang, Chuanming and Gao, Jiaxiang},
  year         = {2022},
  journal      = {IEEE Transactions on Intelligent Transportation Systems},
  volume       = {23},
  number       = {12},
  pages        = {24524--24539},
  doi          = {10.1109/TITS.2022.3210170},
}

@article{alam2021predicting,
  title        = {Predicting irregularities in arrival times for transit buses with recurrent neural networks using GPS coordinates and weather data},
  author       = {Alam, Omar and Kush, Anshuman and Emami, Ali and Pouladzadeh, Parisa},
  year         = {2021},
  journal      = {Journal of Ambient Intelligence and Humanized Computing},
  publisher    = {Springer},
  volume       = {12},
  number       = {7},
  pages        = {7813--7826},
  doi          = {10.1007/s12652-020-02572-8},
}

@article{arrivalNet,
  title        = {ArrivalNet: Predicting City-wide Bus/Tram Arrival Time with Two-dimensional Temporal Variation Modeling},
  author       = {Li, Zirui and Wolf, Patrick and Wang, Meng},
  year         = {2024},
  journal      = {CoRR},
  volume       = {abs/2410.14742},
  doi          = {10.48550/ARXIV.2410.14742},
  url          = {https://doi.org/10.48550/arXiv.2410.14742},
  eprinttype   = {arXiv},
  eprint       = {2410.14742},
  bibsource    = {dblp computer science bibliography, https://dblp.org},
}

@inproceedings{boudabbous2024analyzing,
  title        = {Analyzing public transit schedule deviations: A case study on {Montreal} using real-time data},
  author       = {Boudabbous, Emna and Karaa, Mohamed and Sboui, Lokman and Montecinos, Julio and Alam, Omar},
  year         = {2024},
  booktitle    = {2024 IEEE 27th International Symposium on Real-Time Distributed Computing (ISORC)},
  publisher    = {IEEE},
  pages        = {1--6},
  doi          = {10.1109/ISORC61049.2024.10551354},
  note         = {Exploratory case study on which current systematic methodology builds},
}

@article{chen2025understanding,
  title        = {Understanding bus delay patterns under different temporal and weather conditions: A Bayesian Gaussian mixture model},
  author       = {Chen, Xiaoxu and Saidi, Saeid and Sun, Lijun},
  year         = {2025},
  journal      = {Transportation Research Part C: Emerging Technologies},
  publisher    = {Elsevier},
  volume       = {171},
  pages        = {105000},
  doi          = {10.1016/j.trc.2025.105000},
}

@article{elassy2024intelligent,
  title        = {Intelligent transportation systems for sustainable smart cities},
  author       = {Elassy, Mohamed and Al-Hattab, Mohammed and Takruri, Maen and Badawi, Sufian},
  year         = {2024},
  journal      = {Transportation Engineering},
  publisher    = {Elsevier},
  volume       = {16},
  pages        = {100252},
  doi          = {10.1016/j.treng.2024.100252},
}

@article{elsa2018evaluation,
  title        = {Evaluation of Clustering Algorithms for the Prediction of Trends in Bus Travel Time},
  author       = {Shaji, Hima Elsa and Tangirala, Arun K. and Vanajakshi, Lelitha},
  year         = {2018},
  journal      = {Transportation Research Record: Journal of the Transportation Research Board},
  publisher    = {SAGE Publications},
  volume       = {2672},
  number       = {45},
  pages        = {242--252},
  doi          = {10.1177/0361198118791365},
}

@article{kartini2025dimensionality,
  title        = {Dimensionality reduction using principal component analysis and genetic algorithm for microarray classification},
  author       = {Kartini, D. and others},
  year         = {2025},
  journal      = {Indonesian Journal of Electronics, Electromedical Engineering, and Medical Informatics},
  volume       = {7},
  number       = {1},
  pages        = {23--35},
  doi          = {10.17815/jelectrmed.2025.01},
}

@inproceedings{manual1,
  title        = {Machine Learning for Bus Travel Prediction},
  author       = {Pa{\l}ys, {\L}ukasz and Ganzha, Maria and Paprzycki, Marcin},
  year         = {2022},
  booktitle    = {Computational Science -- ICCS 2022},
  publisher    = {Springer International Publishing},
  address      = {Cham},
  series       = {Lecture Notes in Computer Science},
  volume       = {13351},
  pages        = {703--710},
  doi          = {10.1007/978-3-031-08754-7_72},
  isbn         = {978-3-031-08754-7},
  editor       = {Groen, Derek and de Mulatier, Cl{\'e}lia and Paszynski, Maciej and Krzhizhanovskaya, Valeria V. and Dongarra, Jack J. and Sloot, Peter M. A.},
}

@article{manual2,
  title        = {Modelling bus delay at bus stop},
  author       = {Huo, Yueying and Li, Wenquan and Zhao, Jinhua and Zhu, Shoulin},
  year         = {2018},
  month        = {Jan.},
  journal      = {Transport},
  volume       = {33},
  number       = {1},
  pages        = {12--21},
  doi          = {10.3846/16484142.2014.1003324},
  url          = {https://journals.vilniustech.lt/index.php/Transport/article/view/124},
}

@article{markovic2015analyzing,
  title        = {Analyzing passenger train arrival delays with support vector regression},
  author       = {Markovi{\'c}, Nikola and Milinkovi{\'c}, Sanjin and Tikhonov, Konstantin S. and Schonfeld, Paul},
  year         = {2015},
  journal      = {Transportation Research Part C: Emerging Technologies},
  publisher    = {Elsevier},
  volume       = {56},
  pages        = {251--262},
  doi          = {10.1016/j.trc.2015.03.027},
}

@article{petersen2019multi_output,
  title        = {Multi-output bus travel time prediction with convolutional LSTM neural network},
  author       = {Petersen, Niklas Christoffer and Rodrigues, Filipe and Pereira, Francisco Camara},
  year         = {2019},
  journal      = {Expert Systems with Applications},
  publisher    = {Elsevier},
  volume       = {120},
  pages        = {426--435},
  doi          = {10.1016/j.eswa.2018.11.028},
}

@article{sadegzadeh2024comparative,
  title        = {Comparative analysis of dimensionality reduction techniques: {PCA}, {Laplacian} score, and chi-square on {EEG} classification},
  author       = {Sadegh-Zadeh, S.A. and others},
  year         = {2024},
  journal      = {Scientific Reports},
  volume       = {14},
  pages        = {21456},
  doi          = {10.1038/s41598-024-71234-9},
}

@inproceedings{sc24supercomputing,
  title        = {Neural Network Optimization and Performance Analysis for Real-Time Object Detection at the Edge},
  author       = {Ghose, Animesh and Ren, Yihui and Cui, Yonggang},
  year         = {2024},
  booktitle    = {Proceedings of the SC24 International Conference for High Performance Computing, Networking, Storage and Analysis -- Research Poster Archive},
  address      = {Atlanta, GA, USA},
  pages        = {1--1},
  url          = {https://sc24.supercomputing.org/proceedings/poster/poster\%5Fpages/post172.html},
  note         = {Research poster (ACM Student Research Competition)},
}

@article{shaji2022joint,
  title        = {Joint clustering and prediction approach for travel time prediction},
  author       = {Shaji, Hima Elsa and Tangirala, Arun K. and Vanajakshi, Lelitha},
  year         = {2022},
  journal      = {PLOS ONE},
  publisher    = {Public Library of Science},
  volume       = {17},
  number       = {9},
  pages        = {e0275030},
  doi          = {10.1371/journal.pone.0275030},
}

@article{singh2022review,
  title        = {A review of bus arrival time prediction using artificial intelligence},
  author       = {Singh, Nisha and Kumar, Kranti},
  year         = {2022},
  journal      = {Wiley Interdisciplinary Reviews: Data Mining and Knowledge Discovery},
  publisher    = {Wiley Online Library},
  volume       = {12},
  number       = {4},
  pages        = {e1457},
  doi          = {10.1002/widm.1457},
}

@misc{sun2025realtimebustraveltime,
  title        = {Real-time Bus Travel Time Prediction and Reliability Quantification: A Hybrid Markov Model},
  author       = {Yuran Sun and James Spall and Wai Wong and Xilei Zhao},
  year         = {2025},
  url          = {https://arxiv.org/abs/2503.05907},
  eprint       = {2503.05907},
  archiveprefix = {arXiv},
  primaryclass = {stat.AP},
}

@article{suwardo2010arima,
  title        = {ARIMA models for bus travel time prediction},
  author       = {Suwardo, W. and Napiah, M. and Kamaruddin, I.},
  year         = {2010},
  journal      = {Journal of The Institution of Engineers, Malaysia},
  volume       = {71},
  number       = {2},
  pages        = {49--58},
  note         = {Paper based on data from Ipoh–Lumut corridor, Malaysia},
}

@misc{uber2018h3,
  title        = {H3: A hexagonal hierarchical geospatial indexing system},
  author       = {Brodsky, Isaac},
  year         = {2018},
  note         = {Uber Technologies},
  howpublished = {\url{https://eng.uber.com/h3/}},
}

@article{vlahogianni2014short_term_review,
  title        = {Short-term traffic forecasting: Overview of objectives and methods},
  author       = {Vlahogianni, Eleni I. and Karlaftis, Matthew G. and Golias, John C.},
  year         = {2014},
  journal      = {Transport Reviews},
  publisher    = {Taylor \& Francis},
  volume       = {34},
  number       = {1},
  pages        = {4--24},
  doi          = {10.1080/01441647.2013.951992},
}

@article{xue2025dmtngnn,
  title        = {Data Mining in Transportation Networks with Graph Neural Networks: A Review and Outlook},
  author       = {Xue, Jiawei and Tan, Ruichen and Ma, Jianzhu and Ukkusuri, Satish V.},
  year         = {2025},
  journal      = {CoRR},
  volume       = {abs/2501.16656},
  doi          = {10.48550/arXiv.2501.16656},
  url          = {https://arxiv.org/abs/2501.16656},
  archiveprefix = {arXiv},
  eprint       = {2501.16656},
  primaryclass = {cs.LG},
}

@book{zaharia2018spark_guide,
  title        = {Spark: The Definitive Guide: Big Data Processing Made Simple},
  author       = {Chambers, Bill and Zaharia, Matei},
  year         = {2018},
  publisher    = {O'Reilly Media},
  address      = {Sebastopol, CA},
  isbn         = {978-1-491-92480-6},
}
\end{document}